\documentclass[accepted]{uai2026} 
                        
\usepackage[american]{babel}

\usepackage{hyperref}

\usepackage{amsmath}
\usepackage{amsthm}
\usepackage{booktabs}
\usepackage{algorithm}
\usepackage{algorithmic}
\usepackage[switch]{lineno}
\usepackage{algorithm}
\usepackage{algorithmic}
\usepackage{xcolor}
\usepackage{multirow}
\usepackage{colortbl}
\usepackage{bbding}
\usepackage{subcaption}
\usepackage{soul}
\usepackage{comment}
\usepackage{graphicx}
\usepackage{amsfonts, amsmath}
\usepackage{listings}
\usepackage{bm}
\usepackage{graphicx}
\usepackage{xcolor}
\usepackage{amsmath}
\usepackage{fnpos}
\usepackage{graphicx}
\usepackage{latexsym}
\usepackage{adjustbox}
\usepackage{booktabs}
\usepackage{subfiles}
\usepackage{amsmath}
\usepackage{tabularx}
\usepackage{comment}
\usepackage{siunitx}
\usepackage{pdflscape}
\usepackage{multirow}
\usepackage{soul}
\usepackage{enumitem}
\usepackage{caption}
\usepackage{adjustbox}
\usepackage{sepfootnotes}
\usepackage{longtable}
\usepackage[table]{xcolor}
\usepackage{graphicx}
\usepackage{array}
\usepackage{makecell}

\usepackage{xspace}

\usepackage{parskip}

\theoremstyle{plain}

\theoremstyle{definition}

\theoremstyle{remark}

\newcommand{\xhdr}[1]{\noindent{{\bf #1.}}}

\usepackage{natbib} 
\usepackage{mathtools} 
\usepackage{booktabs} 
\usepackage{tikz} 

\title{Model Agnostic Graph Prompt Learning for Crystal Property Prediction}

\author[1,$\dagger$]{\href{mailto:mcssm2164@iacs.res.in}{Shrimon Mukherjee}}
\author[2,$\dagger$]{\href{mailto:kishalaydas@kgpian.iitkgp.ac.in}{Kishalay Das}}
\author[1]{Partha Basuchowdhuri}
\author[2]{Pawan Goyal}
\author[2]{Niloy Ganguly}
\affil[1]{%
    School of Mathematical \& Computational Sciences\\ 
    Indian Association for the Cultivation of Science\\ 
    Kolkata\\ 
    India
}
\affil[2]{%
    Department of Computer Science and Engineering\\
    Indian Institute of Technology Kharagpur\\
    India
}
\affil[$\dagger$]{%
    Equal Contribution
}
  
\begin{document}
\maketitle

\begin{abstract}
   Graph Neural Networks have emerged as a powerful tool for the fast and accurate prediction of various crystal properties. These models often encode domain-specific knowledge into their graph encoding modules, which increases their parameter size and makes their performance heavily dependent on domain expertise. Added to this, explicitly incorporating all chemical and structural features, that might influence a specific crystal property into the GNN encoder, is a challenging task. In this work, we propose a soft prompt learning framework that captures latent features essential for property prediction, which are not explicitly provided to the GNN. We introduce a novel multilevel graph prompt learning framework comprising both node-level and graph-level soft prompts. At the node level, we capture the local chemical semantics of different atom types, while at the graph level, we encode the global structural symmetry of the crystal graph. Our proposed prompt learning framework is lightweight and seamlessly integrates with any existing GNN encoder. Extensive experiments on popular benchmark datasets show that incorporating prompt learning significantly improves (3\% - 15\%) the performance of state-of-the-art GNN models in crystal property prediction tasks. Furthermore, the learned soft prompts enable cross-property knowledge transfer, enhancing prediction performance for properties with limited training data. Code is available at~\url{https://github.com/shrimonmuke0202/Prompt.git} 
\end{abstract}

\section{Introduction}
Predicting crystal properties remains a significant challenge in materials science. Unlike molecules, which are often represented as regular graphs, crystal materials are modeled as a 3D point cloud of atoms arranged in a well-defined shape with a repeating, orderly structure. Accurate and efficient prediction of crystal properties facilitates the identification of new materials with desired characteristics, reducing the need for extensive physical synthesis or testing and significantly accelerating the discovery process. A natural approach to modeling crystals involves representing them as graphs. As a result, numerous studies have focused on developing Graph Neural Network (GNN) models for crystal property prediction.  Earlier approaches~\cite{xie2018crystal,chen2019graph,louis2020graph,Wolverton2020,schmidt2021crystal} construct a multi-edge graph from the 3D crystal structure and apply GNN model to encode the neighborhood structural information around an atom.\\
Subsequently, numerous studies have proposed various GNN architecture variants that integrate specific domain knowledge into the encoder to enhance crystal representation learning. ALIGNN~\cite{choudhary2021atomistic} incorporates bond angular information among edges to encode many-body interactions, Matformer~\cite{yan2022periodic} is designed to be invariant to periodicity to capture repeating patterns explicitly, and PotNet~\cite{lin2023efficient} directly models inter-atomic potentials based on physics principles. A unifying principle across these models is the assumption that crystal properties, such as formation energy, band gap, total energy, etc. are inherently linked to the chemical semantics of the constituent atoms and the 3D periodic structure. Consequently, predicting these properties necessitates capturing both the local chemical semantics at the node-level and the global 3D structural symmetry at the graph level of the crystal material. 
Thus, over the years, state-of-the-art GNN models have progressively improved in encoding these aspects by employing sophisticated architectures resulting in more accurate property predictions but in turn also increasing the number of parameters, leading to longer training and inference times as well as greater computational overhead.
\\
However, explicitly incorporating {\bf all} chemical and structural features that might influence a specific crystal property into GNN models poses a significant challenge. This process requires specialized domain expertise and may involve features that are not yet fully understood or discovered. 
As a result, the features utilized in these models, while potentially extensive, cannot be deemed exhaustive, which inherently constrains their effectiveness. 
One potential solution is to incorporate latent features that can be learned—both at the local node-level and the global graph level—into the modeling process. {These learned latent features will account for these missing attributes and further enhance the expressive power of both simpler models like CGCNN and advanced models such as ALIGNN, Matformer, and PotNet.}\\
In this work, we propose {\bf soft or embedding-based prompt learning} as a novel approach to learning latent features. Building on the success of pixel-level Visual Prompt (VP) techniques ~\cite{bahng2022exploring,wu2023approximated,wu2022unleashing} and the use of learnable soft prompts in NLP tasks \cite{lester2021power,liu2021p}, we introduce a new multi-level graph prompt learning framework for crystal property prediction. This framework combines both node-level (local) and graph-level (global) soft prompts. At node-level, we learn a set of  $k$ independent vectors, with each vector potentially representing a (latent) chemical feature. The impact of these features on a node in the crystal graph is computed using attention scores derived from those $k$ vectors and their linear projections. At the graph level, we design prompt templates that capture key structural features such as crystal geometry and orientation. 
Our proposed prompt learning approach is efficient, adding only \emph{0.32\%} extra parameters to Matformer, and is compatible with any existing state-of-the-art GNN encoder. 
\\
We perform extensive experiments on two widely used benchmark datasets to evaluate the effectiveness of our proposed prompt learning framework against six state-of-the-art GNN models for the crystal property prediction task. We show that our proposed prompt learning framework results in substantial performance gains for existing GNN-based architectures on diverse sets of properties across these benchmark materials datasets.  In specific, we observe that incorporating such lightweight soft prompts enhances the performance of simpler models like CGCNN by 15.11\%, which makes it competitive to advanced models (more parameters) such as ALIGNN (vanilla). Additionally, even for sophisticated models like ALIGNN, Matformer, and PotNet, we observe performance improvements of 5.52\%, 6.64\%, and 3.15\%, respectively using these lightweight soft prompts. These results highlight the significance and innovation of our proposed framework, which enhances the performance of not only simpler models, but also advanced sophisticated models. Furthermore, we compare prompt learning with several alternative architectural modification strategies and find prompt learning to be both more effective and lightweight.
We further observe cross-property knowledge transfer capability of learned soft prompts, which enhance the performance on properties with sparse/limited training data. Moreover, ablation studies validate the significance of both node-level and graph-level prompt learning in effectively capturing the chemical and structural semantics of crystal materials. We provide the code base in the supplementary material.

\begin{figure}[!htb]%
    \centering
    {\includegraphics[width=\columnwidth]{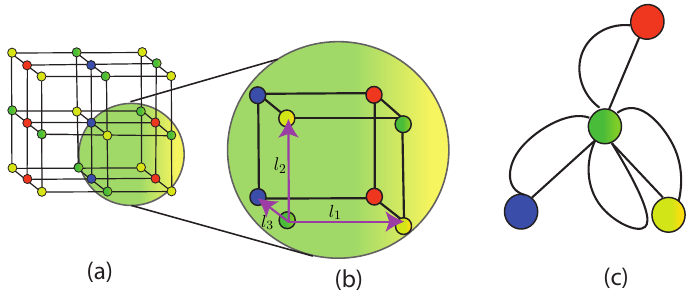}}
    {\caption{(a) A periodic crystal structure, featuring a point cloud of atoms arranged in repeating patterns. (b) A magnified view of a unit cell, clearly delineating the lattice vectors \( L = [l_1, l_2, l_3] \), highlighting the fundamental building blocks of the crystal's geometric arrangement. (c) Multigraph representation of the unit cell.}
    \label{fig:periodic_crystals}}
\end{figure}

\section{Preliminaries}
\subsection{Crystal Representation}
\label{crystal_representation}
Crystal materials are viewed as a 3D point cloud of atoms with an orderly repeating pattern. It can be modeled as a minimal \textit{unit cell}, containing all constituent atoms in specific coordinates and is repeated infinitely in three-dimensional space on a regular lattice, resulting in periodic structures as shown in Fig.~\ref{fig:periodic_crystals}(a) and ~\ref{fig:periodic_crystals}(b). Given a material with $N$ number of atoms in its unit cell, we can describe the unit cell by two matrices: \textit{Atom Type Matrix (\textit{A})} and \textit{Coordinate Matrix ($\textit{C}$)}. Atom Type Matrix $\textit{A} =[\textit{a}_1,\textit{a}_2,...,\textit{a}_N]^T \in \mathbb{R}^{N \times k}$ denotes set of atomic type in one hot representation, where k is the maximum number of possible atom types. Coordinate Matrix $\textit{C}=[\textit{c}_1,\textit{c}_2,...,\textit{c}_N]^T \in \mathbb{R}^{N \times 3}$ denotes atomic coordinate positions, where $\textit{c}_i \in \mathbb{R}^{3}$ corresponds to coordinates of $i^{th}$ atom in the unit cell. Further, there is an additional \textit{Lattice Matrix} $\textit{L}=[\textit{l}_1,\textit{l}_2,\textit{l}_3]^T \in \mathbb{R}^{3 \times 3}$, which describes how a unit cell repeats itself in the 3D space towards $\textit{l}_1,\textit{l}_2$ and $\textit{l}_3$ direction to form the periodic 3D structure of the material. Formally, a given material can be defined as $ \mathbf{\textit{M}}=(\textit{A},\textit{C},\textit{L})$ and we can represent its infinite periodic structure as: $\hat{\textit{C}}  = \{ \hat{c}_i |  \hat{c}_i = \textit{c}_i + \sum_{j=1}^{3} k_j{l}_j \}$ and $\hat{\textit{A}}  = \{ \hat{a}_i |  \hat{a}_i = {a}_i \}$, where $k_1,k_2,k_3, i \in Z, 1 \leq i \leq N$.
\begin{figure*}[!htbp]%
    \centering
    \includegraphics[width=0.80\textwidth]{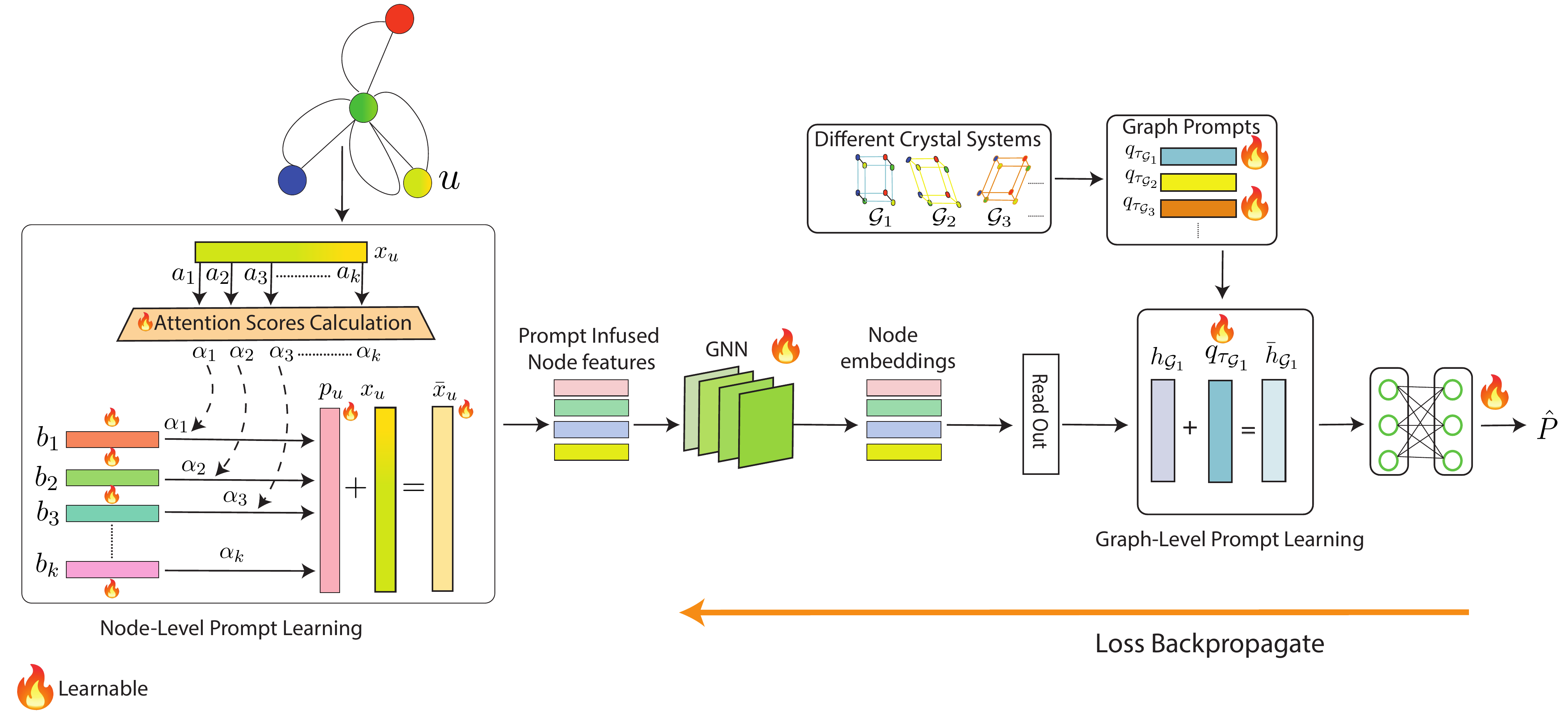}
    {\caption{The overall framework of our universal prompt learning method. For node-level prompting, we use a template that distinguishes between different atom types based on their chemical semantics. At the graph level, our prompt template captures the inherent crystal structural symmetry in the graph representation.}
    \label{fig:model}}
\end{figure*}

\section{Methodology}
Given a property-tagged crystal dataset $\mathcal{D}_{t}=\{(\mathcal{G}_1,y_1), (\mathcal{G}_2,y_2) \dots$ $(\mathcal{G}_m,y_m)\}$ for property prediction, where crystal structure is represented as multi-graphs $\mathcal{G}_i =(\mathcal{V}_i, \mathcal{E}_i, \mathcal{X}_i, \mathcal{F}_i )$, the objective of the property prediction task is to predict $y_i \in \mathbb{R}$ based on the graph representation of the crystal graph $h_{\mathcal{G}_i}$. Formally, this involves learning a mapping function: $f : \mathcal{G}_i \rightarrow y_i$.
The conventional approach to model the aforementioned function $f$ is to use Graph Neural Network (GNN) architectures due to their superior ability to encode graph-structured data. To further improve prediction performance, we incorporate graph prompt learning into GNN models to learn latent chemical and structural semantics tailored to a specific property. Next in this section, we first provide a brief background about the crystal property predictor, followed by the proposed prompt learning framework with detailed explanation of node and graph level prompts, and finally, the learning process. 
\subsection{Crystal Property Predictor}
Conventional GNN based crystal property predictor has two building blocks : (a)  a graph embedding module $f_\theta$ comprising multiple GNN layers, and (b) a projection head $z_\phi$, where 
$\theta$ and $\phi$ are the parameters of the respective modules.
\\
The graph embedding module takes multi-edge graph structure along with node and edge features as input and generates a set of node embeddings $\mathcal{H}$ as output. 
\begin{equation}
    \mathcal{H} = \{h_1,..., h_{|\mathcal{V}|}\}= f_\theta(\mathcal{A}, \mathcal{X}, \mathcal{F})
\end{equation}
where $\mathcal{A}$ is the adjacency matrix of the crystal graph. Further, a READOUT operation (symmetric aggregation e.g. average, sum, or graph pooling) is performed to get the representation $h_{\mathcal{G}}$ for the whole crystal graph $\mathcal{G}$. This graph representation is then input into a projection head $z_\phi$, which predicts the property value $\hat{y}$. 
\begin{equation}
\begin{split}
     \hat{y} = z_\phi ( h_{\mathcal{G}}) ;  \:\:\: h_{\mathcal{G}} = READOUT\biggl\{  h_u^{(L)} | u \in  \mathcal{V}   \biggr\}
\end{split}
\end{equation}
The entire crystal property predictor architecture, including the GNN layers and the projection head, is trained end-to-end to optimize their parameters on downstream property prediction tasks.
\subsection{Crystal Prompt Learning}
\label{crys_prompt}
The effectiveness of a GNN-based crystal property predictor largely depends on its ability to encode the chemical semantics of the constituent atoms and the crystal structural topology into a lower-dimensional embedding space, thereby generating the graph representation $h_{\mathcal{G}}$. However, explicitly providing all the chemical and structural semantics that influence a specific crystal property to GNN models is inherently challenging. 
Consequently, certain latent 
features, both at the local node-level and the global graph level, that affect specific crystal property, are often not captured in the existing models.

To capture the impact of such latent features, we propose a \emph{Multi-level Graph Prompt Learning} approach (Fig.~\ref{fig:model}), introducing learnable soft prompts at both the node and graph levels to capture essential chemical and structural latent features for specific crystal property prediction. Recently, soft prompts have been effectively utilized to integrate additional meta-information into input data across various computer vision and NLP tasks. In computer vision, pixel-level Visual Prompt (VP) techniques \cite{bahng2022exploring,wu2022unleashing,wu2023approximated} modify input images by adding learnable soft visual prompts, which serve as visual cues to guide the decision-making of machine learning models. Similarly, in NLP, researchers \cite{liu2021p,lester2021power} have demonstrated significant potential in incorporating additional meta-information by introducing learnable soft prompts into sentences, such as adding positional encodings to input tokens to convey positional information.

Building on these recent successes, in our proposed methodology, we transformed the input graph $\mathcal{G}$ to prompted graph $\psi{(\mathcal{G})}$ by introducing learnable soft prompts into both the feature space (input node feature) and latent space (graph embedding) of the crystal graph. 
We fed the transformed graph $\psi{(\mathcal{G})}$ into a GNN model and trained the learnable soft prompts together with the GNN model parameters in an end-to-end supervised manner. In the following subsections, we provide a detailed description of both the node and graph-level prompting methods.
\subsubsection{Node-Level Prompt}
\label{sec_node_prompt}
Goal of node/atom level prompt learning is to capture untapped  chemical attributes (local properties), 
which play a crucial role in predicting a particular crystal property. 
To this effect we extend the idea of the universal graph prompt technique proposed by~\cite{fang2024universal} and
posit that there are $k$ different chemical features ($k$ is a hyper-parameter) which we learn through $k$ independent vectors: $b_1,b_2,..b_k \in \mathbb{R}^d$ and $k$ linear projections $a_1,a_2,..a_k$. 
The learned vectors are utilized through an attention-based aggregation mechanism to produce node-level prompts.\\
First, for each node $u$, we compute $k$ attention scores leveraging softmax over linear projection of node feature $x_u$:
\begin{equation}
\label{eq:node_prompt_attn}
    \alpha_{u,j} = \frac{exp(a^T_jx_u)}{\sum^k_l exp(a^T_lx_u)}
\end{equation}
Further, we compute prompt vector $p_u$ through attention-based aggregation leveraging attention scores and those $k$ vectors:
\begin{equation}
\label{eq:node_prompt_adv}
    p_{u} = \sum^k_j \alpha_{u,j} b_j 
\end{equation}
We apply the same process to each node, generating a unique prompt vector for each one. These prompt vectors are then combined with their corresponding input node features to create the prompted node feature set $\mathcal{X}'$
\begin{equation}
\label{eq:node_prompt}
    \mathcal{X'}=\{\bar{x}_u = x_u + p_u | u \in \mathcal{V})\}
\end{equation}
This prompted node feature set $\mathcal{X}'$ is subsequently fed into the GNN-based graph embedding module to generate the graph representation $h_\mathcal{G}$ for the entire crystal graph $\mathcal{G}$.
\begin{equation}
\begin{split}
    \mathcal{H} = f_\theta (\mathcal{A}, \mathcal{X}', \mathcal{F}); \quad
     h_{\mathcal{G}} = READOUT\{ \mathcal{H} \}
\end{split}
\end{equation}
These \(k\) vectors $b_i$ and projection layers $a_i$ are trained on downstream property prediction task (Equation \ref{eq:mse_loss}) for a specific property of interest. Intuitively, these \(k\) vectors represent \(k\) latent chemical features essential for the property but not explicitly provided to the GNN through the input node features. Additionally, the projection layers are trained to determine the importance (attention) of these latent features for a specific node based on its corresponding node features. The hyperparameter \(k\) was found to produce optimal results when set to 10 based on empirical testing. As a result, the additional parameter overhead to the GNN model is minimal. 
\subsubsection{Graph-Level Prompt}
In addition to the atomic chemical semantics, crystal properties are strongly correlated with the geometric symmetry of crystal structures. By definition, a crystal material is a periodic arrangement of repeating “motifs”( e.g. atoms, ions). The space group of a crystal describes the symmetry of that crystal which is an important aspect of that crystal's internal structure. In turn crystals across certain space groups show similarities among each other, hence they are divided among seven crystal systems: Triclinic, Monoclinic, Orthorhombic, Tetragonal, Trigonal, Hexagonal, and Cubic system ~\cite{kittel2018kittel}.\\
The goal of graph-level prompt learning is to design a prompt template that captures underlying crystal geometry in graph representation. Toward that goal, we train distinct soft prompts for each crystal system and combine them with the graph embeddings. In specific, we learn an embedding matrix $\mathbf{Q} \in \mathbb{R}^{7 \times d}$, containing learnable soft prompts corresponding to each crystal system. For any specific, crystal graph $\mathcal{G}$, with crystal system $\tau_\mathcal{G}$, we retrieve $\tau_\mathcal{G}$ row of $\mathbf{Q}$ and add it as soft prompts $q_{\tau_\mathcal{G}}$ with the graph embedding $h_{\mathcal{G}}$. Formally,
\begin{equation}
\begin{split}
\label{eq:graph_prompt}
    q_{\tau_\mathcal{G}} \leftarrow \mathbf{Q}[\tau_\mathcal{G},:]; \quad
    \bar{h}_{\mathcal{G}} = h_{\mathcal{G}} + q_{\tau_\mathcal{G}}
\end{split}
\end{equation}
The updated graph representation, $\bar{h}_{\mathcal{G}}$, is subsequently passed through a projection head, $z_\phi$, to predict the property values.
\subsubsection{Prompt Learning Framework} 
For a specific property prediction task, we convert the input crystal graph into a prompted graph by integrating node-level prompts with the input node features and graph-level prompts with the graph embeddings. This prompted graph is then fed into a state-of-the-art crystal GNN model, which is trained end-to-end. We use mean square error (MSE) objective function between predicted $\hat{\mathcal{Y}}$ and true property values $\mathcal{Y}$ 
\begin{equation}
\label{eq:mse_loss}
                \underset{\theta,\phi,b^*_i,a^*_i,q_{\tau_\mathcal{G}}}{min} \ \mathcal{L}_{MSE}=  \lVert {\hat{\mathcal{Y}}} - {\mathcal{Y}} \rVert^2
\end{equation}
where $b^*_i$ and $a^*_i$ are the independent vectors and linear projections for node-level prompts respectively, $q_{\tau_\mathcal{G}}$ is the graph-level prompt and $\theta$ \& $\phi$ are the trainable parameters of GNN module and projection head respectively.\\
The proposed prompt learning framework is 
trained from scratch and optimized for any downstream property prediction task. All these independent vectors $b^*_i$, projection layers $a^*_i$ and graph level prompts $q_{\tau_\mathcal{G}}$ are randomly initialized and trained along with other model parameters of the GNN property predictor. We hypothesize that training enables these soft prompt templates to capture intrinsic chemical and structural semantics, which is essential for estimating various crystal properties.
\begin{table*}
\centering
\setlength{\tabcolsep}{9 pt}
\resizebox{1.0\textwidth}{!}{
\begin{tabular}{c | c c | c c| c c | c c | c c | c c}
         \toprule  
         Property & CGCNN & CGCNN & ALIGNN & ALIGNN & Matformer & Matformer & PotNet & PotNet & iComFormer & iComFormer & eComFormer & eComFormer\\
         & & (Prompt) & & (Prompt) & & (Prompt) & & (Prompt) & & (Prompt) &  & (Prompt)\\
         \midrule
          \#Parameter &  0.423M & 0.433M & 1.10M & 1.11M & 3.27M &  3.28M & 0.473M & 0.484M & 1.38M & 1.39M  & 3.72M & 3.73M \\
          \midrule
          Formation Energy & 0.063 & \textbf{0.041} & 0.035 & \textbf{0.033} & 0.033 & \textbf{0.032} & 0.0294 & \textbf{0.0292} & 0.027 & \textbf{0.026}
          & 0.028 & \textbf{0.027}\\
          
         Bandgap(OPT) & 0.200 & \textbf{0.155} & 0.142 & \textbf{0.140} & 0.137 & \textbf{0.136} & 0.1274 & \textbf{0.1270} & 0.135 & \textbf{0.134} & 0.124 & 
         \textbf{0.122}\\
         
         Total Energy & 0.078 & \textbf{0.047} & 0.037 & \textbf{0.035} & 0.035 & \textbf{0.033} & 0.0324 & \textbf{0.0311} & 0.031 & \textbf{0.029} & 0.032 & 
         \textbf{0.031}\\
         
         Ehull & 0.170 & \textbf{0.116} & 0.076 & \textbf{0.064} & 0.064 & \textbf{0.053} & 0.0552 & \textbf{0.0481} & 0.047 & \textbf{0.046} & 0.044 & \textbf{0.040} \\
         
         Bandgap(MBJ) & 0.410 & \textbf{0.354} & 0.310 & \textbf{0.299} & 0.300 & \textbf{0.285} & 0.2701 & \textbf{0.2580} & 0.264 & \textbf{0.259} & 0.280 & \textbf{0.276}\\
         
         Bulk Modulus(Kv) & 14.47 & \textbf{12.54} & 10.40 & \textbf{10.21} & 11.21 & \textbf{10.05} & 10.110 & \textbf{9.9810} & 14.27 & \textbf{13.22} & 13.45 & \textbf{12.26}\\
         
         Shear Modulus(Gv) & 11.75 & \textbf{10.58} & 9.862 & \textbf{9.731} & 10.76 & \textbf{9.581} & 9.2320 & \textbf{9.0201} & 9.760 & \textbf{9.359} & 11.12 & \textbf{10.56}\\
         
         SLME(\%) &5.660 & \textbf{4.801} & 4.521 & \textbf{4.420} & 5.260 & \textbf{4.764} & 4.5701 & \textbf{4.4310} & 4.702 & \textbf{4.342} & 4.662 & \textbf{4.650}\\
         
         Spillage & 0.397 & \textbf{0.388} & 0.351 & \textbf{0.348} & 0.398 & \textbf{0.359} & 0.3610 & \textbf{0.3522} & 0.369 & \textbf{0.351} & 0.382 & \textbf{0.369}\\
         \midrule
         
         Formation Energy & 0.031 & \textbf{0.029} & 0.022 & \textbf{0.021} & 0.021 & \textbf{0.020} & 0.0190 & \textbf{0.0183} & 0.018 & \textbf{0.017} & 0.018 & \textbf{0.016}\\
         
         Bandgap(OPT) & 0.292 & \textbf{0.288} & \textbf{0.218} & 0.223 & 0.211 & \textbf{0.210} & 0.2041 & \textbf{0.1942} & 0.193 & \textbf{0.190} & 0.202 & \textbf{0.199}\\
         
         Bulk Modulus(Kv) & 0.047 & \textbf{0.043} & 0.051 & \textbf{0.038} & 0.043 & \textbf{0.042} & 0.0400 & \textbf{0.0382} & 0.038 & \textbf{0.037} & 0.042 & \textbf{0.039}\\
         
         Shear Modulus(Gv) & \textbf{0.077} & \textbf{0.077} & 0.078 & \textbf{0.069} & 0.073 & \textbf{0.067} & 0.0654 & \textbf{0.0650} & 0.064 & \textbf{0.063} & 0.073 & \textbf{0.072}\\
        \bottomrule     
    \end{tabular}}
    \caption{Summary of the results (MAE) of different properties in JARVIS-DFT (top) and 
Materials Project (bottom). Model M is the vanilla variant of the SOTA model and M (Prompt) is a prompt-infused variant (both node- and graph-level) of the SOTA model. The best performance has been highlighted in \textbf{bold}. }
\label{tab:main_result}
\end{table*}

\section{Experimental Setup}
To assess the effectiveness of our proposed graph prompt learning framework, we perform experiments using two benchmark crystal datasets: Materials Project~\cite{MP} and JARVIS-DFT~\cite{choudhary2020joint}. We choose {\bf six popular state-of-the-art GNN models} for supervised crystal property prediction; these are CGCNN~\cite{xie2018crystal}, ALIGNN~\cite{choudhary2021atomistic}, Matformer~\cite{yan2022periodic}, PotNet~\cite{lin2023efficient}, iComFormer~\cite{yan2024complete}, and eComFormer~\cite{yan2024complete}. For a thorough evaluation, we examine both the vanilla variant and the prompt-infused variant of these SOTA models (Section~\ref{sec_downstream_result}). 
To implement prompt-infused variants, we used the available official PyTorch implementations of all the baselines and incorporated node- and graph-level prompts into it.
We train each prompt-infused model for 1000 epochs using AdamW~\cite{loshchilov2017decoupled} optimizer with normalized weight decay of $10^{-5}$ and keep the batch size as 64. We schedule the learning rate according to the one-cycle policy~\cite{smith2018disciplined} with a maximum learning rate of 0.001. We keep our atom-type prompt dimension the same as the embedding dimension of the atom and for graph-level prompt we keep the dimension the same as the embedding dimension of the graphs. We perform all the experiments on shared servers with NVIDIA A6000 GPUs.

\section{Experimental Results}
\label{sec_downstream_result}
In this section, we first evaluate the performance of our proposed prompt learning framework against baseline models using benchmark datasets and provide a detailed analysis (Section \ref{sec_benchmark_result}). Further, we asses the performance of the prompt learning model in conjunction with an existing pre-trained model and compare it against well-established methods, including graph prompt tuning and conventional fine-tuning approaches (Section \ref{appendix_pretrain}). Next, we compare the effectiveness of prompt learning with alternative architectural modifications (Section \ref{arch_modification}). Additionally, we demonstrate the cross-property knowledge transfer capability of learned prompts in limited data scenarios (Section \ref{cross_prop}). Finally, we conduct ablation studies to analyze the impact of different prompt modalities (Section~\ref{ablation_prompt_modality}) and sensitivity on the number of basis vectors (Section~\ref{appendix_basis}).
\begin{table*}[!htb]
 \centering
    \setlength{\tabcolsep}{2.2 pt}
    \resizebox{\textwidth}{!}{
     \begin{tabular}{c | c c c c c| c c c c c | c c c c c}
         \toprule  
         Property & CrysGNN & CrysGNN & CrysGNN & CrysGNN & CrysGNN & Crystal & Crystal Twins & Crystal Twins & Crystal Twins & Crystal Twins & DPF & DPF & DPF & DPF & DPF\\
         & & (GPF) & (SUPT) & (LEAP) & (Prompt) & Twins & (GPF) & (SUPT) & (LEAP) & (Prompt) & & (GPF) & (SUPT) & (LEAP) & (Prompt)\\
         \midrule
         Formation Energy & 0.146 & 0.738 & 0.222 & 0.322 & \textbf{0.133} & 0.044 & 0.245 & 0.199 & 0.322 & \textbf{0.041} & 0.029 & 0.099 & 0.148 & 0.184 & \textbf{0.027} \\
         
         Bandgap(OPT) & 0.325 & 0.810 & 0.426 & 0.520 & \textbf{0.310} & 0.164 & 0.480 & 0.392 & 0.536 & \textbf{0.151} & 0.123 & 0.277 & 0.327 & 0.401 & \textbf{0.120} \\
         
         Total Energy & 0.196 & 1.503 & 0.479 & 0.729 & \textbf{0.164} & 0.051 & 0.399 & 0.255 & 0.366 & \textbf{0.045} & 0.032 & 0.104 & 0.182 & 0.221 & \textbf{0.029} \\
         
         Ehull & 0.227 & 0.979 & 0.439 & 0.725 & \textbf{0.205} & 0.134 & 0.364 & 0.283 & 0.637 & \textbf{0.119} & 0.063 & 0.180 & 0.432 & 0.125 & \textbf{0.054} \\
         
         Bandgap(MBJ) & 0.601 & 1.351 & 0.712 & 0.999 & \textbf{0.587} & 0.379 & 0.849 & 0.695 & 0.996 & \textbf{0.337} & 0.316 & 0.529 & 0.590 & 0.625 & \textbf{0.287}\\
         
         Bulk Modulus(Kv) & 17.37 & 49.49  & 19.36 & 35.23 & \textbf{16.79} & 13.08 & 14.83  & 18.77 & 29.39 & \textbf{12.44} & 10.43 & 16.57 & 18.99 & 12.19 & \textbf{10.30}\\
         
         Shear Modulus(Gv) & 13.19 & 25.21 & 14.02 & 21.22 & \textbf{13.12} & 10.55 & 19.92 & 14.88 & 20.21 & \textbf{10.16} & 9.596 & 13.62 & 14.17 & 10.35 & \textbf{9.498}\\
         
         SLME(\%) & 6.192 & 10.45  & 6.403 & 9.253 & \textbf{6.050} & 4.891 & 7.420  & 6.649 & 8.229 & \textbf{4.590} & 5.129 & 6.428 & 6.786 & 5.745 & \textbf{4.724}\\
         
         Spillage & 0.419 & 0.507 & 0.442 & 0.493 & \textbf{0.418} & 0.386 & 0.432 & 0.415 & 0.472 & \textbf{0.383} & 0.358 & 0.425 & 0.436 & 0.518 & \textbf{0.347}\\
         
        \bottomrule     
    \end{tabular}}
    \caption{The performance between finetuned and the prompt-infused versions of the existing pre-trained models on nine materials properties from JARVIS-DFT dataset. The best performance has been highlighted in \textbf{bold}.}
\label{tab:pretrained_models_comp}
\end{table*}

\subsection{Performance on the benchmark datasets}
\label{sec_benchmark_result}
We choose six SOTA models for supervised crystal property prediction, CGCNN, ALIGNN, Matformer, PotNet, iComFormer, and eComFormer, where we compare the performance of the vanilla variant (version reported in the respective papers) of each SOTA model with our proposed prompt-infused variant. To avoid any deterioration of the performance due to insufficient hyperparameter tuning, we report the property prediction results for vanilla variants from the respective papers.
\subsubsection{JARVIS-DFT Dataset.} 
We first conduct experiments on the JARVIS-DFT dataset, which is a widely used large-scale material benchmark containing 55,722 crystals. JARVIS (Joint Automated Repository for Various Integrated Simulations)\footnote{https://jarvis.nist.gov/} is a free and publicly available database that includes crystal structures and corresponding properties from DFT-based calculations, classical force-fields and machine learning approaches~\cite{choudhary2020joint}. In line with previous state-of-the-art studies, we select nine crystal properties for the downstream property prediction task: formation energy, bandgap (OPT), bandgap (MBJ), total energy, bulk moduli, shear moduli, ehull, spillage, and SLME(\%). We use 80\%, 10\%, and 10\% train, validation and test split for all the properties as used by PotNet. We report the mean absolute error (MAE) of the predicted and actual value of a particular property for test data in Table~\ref{tab:main_result} to compare the performance of different participating methods.  We observe that our proposed prompt infused variant of any SOTA model outperforms the vanilla model across all the properties. In specific, the average improvement in CGCNN, ALIGNN, Matformer, PotNet, iComFormer, and eComFormer are 20.37\%, 4.23\%, 8.02\%, 3.49\%, 4.32\%, and 4.04\%  respectively. 
\subsubsection{Materials Project (MP) Dataset.} We further use another benchmark material dataset, Materials Project-2018.6.1, which comprises 69,239 materials. Materials Project is also a free and publicly available database that contains crystal structures and the corresponding calculated materials properties~\cite{MP}. The dataset is created by taking the results obtained with density functional theory (DFT) based calculations. The dataset consists of data on electronic structure, as well as thermodynamic, mechanical, and dielectric characteristics. Additionally, it offers a graphical web-based search platform\footnote{https://materialsproject.org/}.\\
Following the previous SOTA algorithms, here we choose four crystal properties, namely, formation energy, bandgap (OPT), bulk modulus (Kv), and shear modulus (Gv), respectively. We follow the previous line of work to split the data for training, validation, and test. For formation energy and bandgap(OPT), we use 60,000, 5,000, and 4,239 crystals as train, validation, and test split as used by ALIGNN, whereas we use 4,664, 393, and 393 crystals as train, validation, and test split for bulk and shear moduli as used by Matformer. 
In Table~\ref{tab:main_result}, we report the MAE of different baseline models and their prompt-infused variants on the MP dataset. To ensure consistency with the results reported by baseline studies, we present bulk and shear moduli values in GPa for JARVIS-DFT and log(GPa) values for MP. Our findings indicate that the prompt-infused variant of any SOTA model consistently outperforms the standard model across nearly all properties in the MP dataset. In specific, the average improvement in CGCNN, ALIGNN, Matformer, PotNet, iComFormer, and eComFormer are 4.08\%, 9.82\%, 3.94\%, 3.41\%, 2.83\%, and 5.28\% respectively. Additionally, the impressive performance of all the prompt infused variants on the Bulk Moduli and Shear Moduli tasks with just 4,664 training samples highlights the robustness and adaptability of our prompt learning framework in scenarios with limited training data.
Overall, the performance improvements observed across both datasets and various SOTA GNN models demonstrate the effectiveness of our proposed prompt learning framework in capturing latent chemical and structural semantics, thereby generating more enriched crystal representations. 
Historically, the research community has focused on developing more complex GNN architectures to integrate intrinsic domain knowledge into the encoder, resulting in a significant increase in the number of parameters (see the \#Parameters row in Table \ref{tab:main_result}). In contrast, our proposed prompt-tuning framework introduces approximately \emph{10k} additional parameters across all models as node- and graph-level soft prompts. This addition is negligible but substantially improves the performance of simpler models like CGCNN, making them competitive with more advanced models such as ALIGNN, Matformer, PotNet, iComFormer, and eComFormer.
\subsection{Comparison with Pre-trained Models}
\label{appendix_pretrain}
In this section, we investigate whether our proposed prompt learning framework helps us to enhance existing crystal pre-trained models in property prediction performance. Specifically, we compare the prompt-infused variants of these pre-trained models against both conventional graph prompt learning approaches and traditional fine-tuning methods applied to the same crystal pre-trained models. We take CrysGNN~\cite{das2023crysgnn}, Crystal Twins~\cite{magar2022crystal}, and DPF~\cite{shen2025denoising} as the pre-trained models. CrysGNN and Crystal Twins are initially pre-trained using a self-supervised framework, whereas DPF leverages diffusion-based pre-training on a large corpus of property-unlabeled data before being fine-tuned for downstream property prediction tasks. Note that the pre-training frameworks of CrysGNN and Crystal Twins are built upon CGCNN, whereas DPF adopts Matformer as its underlying graph embedding module. We adopt GPF~\cite{fang2024universal}, SUPT~\cite{lee2024subgraph}, and LEAP~\cite{xu2025learning} as state-of-the-art graph prompt learning baselines, where the pre-trained models are kept frozen and only graph prompts are tuned. For our prompt-infused variants, we begin with a crystal pre-trained model, integrate both node-level and graph-level prompts, and subsequently fine-tune the model, along with soft prompts, on a specific crystal property prediction task.
\begin{table}[!htb]
\renewcommand\arraystretch{0.9}
\setlength{\tabcolsep}{2.5pt}
\centering
\resizebox{\columnwidth}{!}{
\begin{tabular}{clcccccp{0.1cm}<{\centering}p{0.1cm}<{\centering}p{0.1cm}<{\centering}p{0.1cm}<{\centering}p{0.1cm}<{\centering}p{0.1cm}<{\centering}p{0.1cm}<{\centering}p{0.1cm}<{\centering}cc<{\centering}cc} \toprule
\makecell{GNN \\ Model} & \makecell{Setup \\ {}} & \makecell{\# Parameters \\ {}} & \makecell{Formation \\ Energy} & \makecell{Bandgap \\ (OPT)} & \makecell{Total \\ Energy} & \makecell{Ehull\\{}}  
\\ 
\midrule

\multirow{6}{*}{} &  Vanilla & 0.423M & 0.063 & 0.200 & 0.078 & 0.170 \\
~ & 2 $\times$ GNN Layers & 0.788M & 0.061 & 0.174 & 0.073 & 0.162\\

~ &  Increase Hidden Dim & 1.435M & 0.055 & 0.177 & 0.072 & 0.175\\

CGCNN & Skip Connection & 0.423M & 0.059 & 0.192 & 0.073 & 0.161\\

~ & Adapter Learning & 0.476M & 0.151 & 0.224 & 0.278 & 0.261\\

~ & Prompt Learning & 0.433M & \textbf{0.041} & \textbf{0.155} & \textbf{0.047} & \textbf{0.116}\\

\midrule

\multirow{5}{*}{} & Vanilla & 3.27M & 0.033 & 0.137 & 0.035 & 0.064\\
~ & 2 $\times$ GNN Layers & 6.34M & 0.032 & 0.143 & 0.035 & 0.057\\

~ &  Increase Hidden Dim & 12.84M & 0.052 & 0.139 & 0.092 & 0.104\\

Matformer & Skip Connection & 3.27M & 0.033 & 0.138 & 0.034 & 0.064\\

~ & Adapter Learning & 3.36M & 0.089 & 0.156 & 0.194 & 0.155\\

~ & Prompt Learning & 3.28M & \textbf{0.032} & \textbf{0.136} & \textbf{0.033} & \textbf{0.053}\\

\bottomrule
\end{tabular}
}
\caption{Summary of results (MAE) for various properties in JARVIS-DFT using different architectural modifications in CGCNN and Matformer. Our proposed prompt learning paradigm is both lightweight and delivers superior performance compared to alternative approaches.}
\label{tab:Different_modalities_new}
\end{table}\\
We report the results of all competing models for nine properties formation energy, bandgap(OPT), total energy, ehull, bandgap(MBJ), bulk moduli(Kv), shear moduli(Gv), SLME, and spillage from JARVIS dataset in Table~\ref {tab:pretrained_models_comp}.  We observe that the prompt-tuned variants of these pre-trained models outperform both conventional graph prompt learning as well as fine-tuned versions by a good margin for all nine properties. Specifically, compared to fine-tuned versions the average improvements in CrysGNN, Crystal Twins, DPF are 5.36\%, 7.14\%, and 6.13\% respectively. This demonstrates that incorporating prompts into pre-trained models and fine-tuning enhances the model's predictive performance compared to the traditional fine-tuning framework. However, when employing a conventional graph prompt learning framework—while keeping the pre-trained model weights frozen—we observe negative transfer and a noticeable degradation in performance across all evaluated properties. One possible reason is that, unlike other graph domains such as molecular structures, the pre-trained models for crystal structures are less powerful and have been trained on relatively limited data. Additionally, existing graph prompt-tuning methods tend to overlook unique characteristics inherent to crystal structures, such as crystal symmetry and other global features. Finally, prior prompt-based methods such as GPF, SUPT, and LEAP have primarily been evaluated on binary classification tasks in molecular datasets, whereas our work focuses on regression-based crystal property prediction, which poses a more challenging learning problem.
\begin{table}[htbp]
\centering
\setlength{\arrayrulewidth}{0.1mm}
\setlength{\tabcolsep}{7pt}
\renewcommand{\arraystretch}{1.5}
\resizebox{1.0\columnwidth}{!}{\begin{tabular}{ccc|ccc}
 \multicolumn{3}{c}{} & \multicolumn{3}{c}{\underline{Prompt Transfer}}  \\
Properties & Vanilla & \makecell{Prompt \\ Learning} & \makecell{Formation \\ Energy} & \makecell{Bandgap \\ (OPT)}  & \makecell{Total \\ Energy} \\
\hline
bg(MBJ) & 0.410 & 0.354 &  \colorbox{green!50}{0.337} & \colorbox{green!20}{0.352} & \colorbox{green!30}{0.340}\\ 
\hline
Shear & 11.75 & 10.58 & \colorbox{red!40}{10.68} & \colorbox{green!50}{10.38} & \colorbox{green!20}{10.40}\\
\hline
Bulk & 14.47 & 12.54 & \colorbox{green!50}{12.06} & \colorbox{green!28}{12.43} & \colorbox{green!30}{12.42}\\ 
\hline
SLME & 5.66 & 4.81 & \colorbox{green!50}{4.68} & \colorbox{red!40}{4.94} & \colorbox{green!30}{4.70} \\ 
\hline
Spillage & 0.397 & 0.388 & \colorbox{green!28}{0.384} & \colorbox{green!30}{0.383} & \colorbox{green!50}{0.380} \\ 
\bottomrule
\end{tabular}}
\caption{Results showing the knowledge transfer capability of soft prompts, trained on properties with more training data and applied to properties with limited training data. }
\label{tab:know_ledge_transfer}
\end{table}

\subsection{Alternate Architectural Modifications}
\label{arch_modification}
Next, we explore alternative methods for learning latent chemical and structural features by making simple modifications to the existing GNN models. The approaches investigated are increasing hidden layer dimensions, doubling the number of GNN layers, introducing skip connections, and implementing adapter learning.
\begin{enumerate}[nolistsep]
    \item \textbf{2 $\times$ GNN Layers.} In this variant, the number of GNN layers is increased by doubling the original number of layers.
    \item \textbf{Increasing Hidden Dim.} In this variant, the hidden layer dimensions for each GNN are increased by 64.
    \item \textbf{Skip Connection.} In this variant, residual connections are added to each layer of the GNN.
    \item \textbf{Adapter Learning.} In this variant, a parameter-efficient adapter-based tuning technique is used for each layer of the GNN. The method follows the approach proposed in AdapterGNN~\cite{li2024adaptergnn}, where parameter-efficient adapters are added to each layer of the GNN models.
\end{enumerate}
We conduct another set of experiments using JARVIS-DFT dataset on CGCNN and Matformer to compare the performance of these alternative methods with prompt learning and report the results in Table~\ref{tab:Different_modalities_new} and Table~\ref{tab:Different_modalities_appendix}. We observe that compared to all other approaches, prompt learning introduces minimal parameter overhead while outperforming all of these. In particular, increasing hidden layer dimensions or doubling the number of GNN layers substantially increases the model's parameter count, yet results in minimal performance improvement across all properties. On the other hand, while adapter learning and skip connections add only a small parameter overhead, they also fail to significantly enhance crystal property prediction performance.
This highlights the ingenuity of our graph prompt learning approach, which is lightweight, model-agnostic, and provide substantial improvement over the vanilla models.
\subsection{Cross-Property Knowledge Transfer Capability of Soft Prompts}
\label{cross_prop}
A major challenge in developing crystal property predictor is the lack of labeled datasets for all properties. For instance, in the JARVIS DFT dataset, properties such as formation energy or bandgap (OPT) typically have larger training data, while properties such as bulk moduli, shear moduli, or bandgap (MBJ) have limited data available. This raises an interesting question: \emph{Can the knowledge captured by soft prompts for well-represented properties be transferred to improve performance on under-represented properties with limited data?} To explore this further, we designed an additional experiment on JARVIS-DFT dataset where soft prompts (both node-level and graph-level) are trained on one of the properties with abundant training data like formation energy, bandgap(OPT), or total energy. While training other properties having limited data, instead of randomly initializing the soft prompts, we initialize using the learned prompts. We report the results of four properties in Table~\ref{tab:know_ledge_transfer}. We observe that when a property with limited training data is correlated with a property that has abundant training data, the soft prompts learned from the latter can improve the performance of the former. For instance, properties like bandgap (MBJ), bulk modulus, and SLME are known to correlate with energies such as formation energy or total energy. Therefore, we observe that transferring knowledge from soft prompts learned on formation energy or total energy further enhances the performance of bandgap (MBJ), bulk modulus, and SLME compared to the conventional prompt learning approach. In contrast, shear modulus is unrelated to formation energy, and SLME is not correlated with bandgap (OPT), leading to slight degradation in performance. Overall, these results highlight the knowledge transfer capabilities of learned soft prompts and enhance model performance in limited data settings.
\begin{figure}[!htb]%
    \centering
    {\includegraphics[width=\columnwidth]{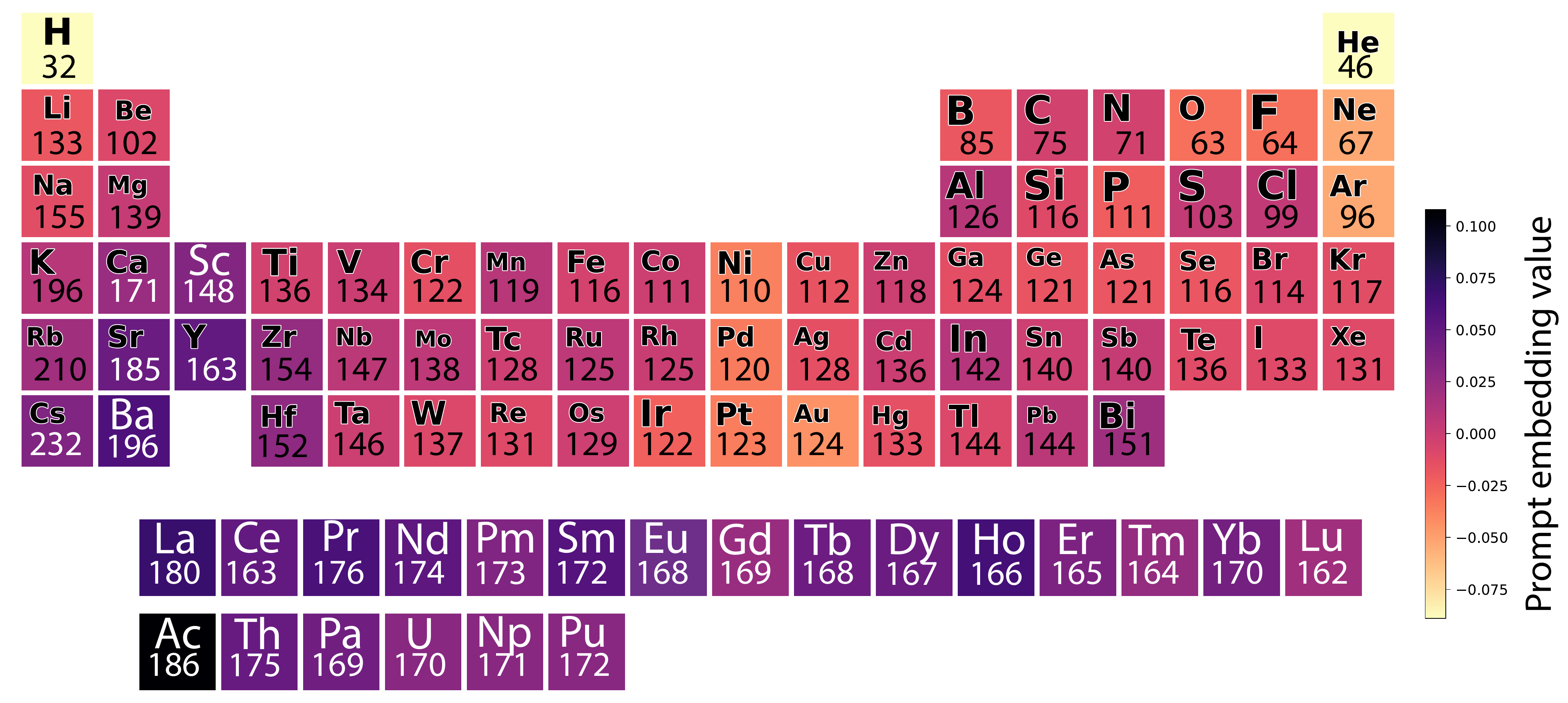}}
    {\caption{Periodic table visualization of the mean node-level prompt embedding values for dimension 7 across elements. Tile colors denote prompt values, while the numbers indicate the original covalent radius (pm).}
    \label{fig:periodic_crystals_radius_covalent}}
\end{figure}

\subsection{Expressiveness of Node-Level Soft Prompt}
\label{softprompt_latent}
To evaluate the expressiveness of the learned node-level soft prompts, we examine whether individual prompt dimensions capture meaningful periodic trends across elements. For each element, we compute the mean value of the corresponding prompt embedding across all occurrences in the dataset and visualize the resulting distribution over the periodic table. Interestingly, we observe that certain prompt dimensions align with chemically meaningful elemental properties. As shown in Fig.~\ref{fig:periodic_crystals_radius_covalent}, prompt embedding dimension seven exhibits clear periodic patterns across elements and shows a strong positive correlation with covalent radius ($r=$0.8095). In particular, electropositive elements with large covalent radii—such as alkali metals (e.g., K, Rb, Cs), alkaline-earth metals (e.g., Ca, Sr, Ba), and lanthanides—tend to exhibit relatively higher prompt values, whereas smaller non-metallic elements such as C, N, O, and F generally show lower values. This pattern broadly follows the periodic variation of covalent radius~\cite{cordero2008covalent,pyykk02015additive}, suggesting that the learned prompt embeddings implicitly encode atomic-size-related information relevant to bonding environments in crystalline materials. Additional prompt dimensions also show correlations with other elemental descriptors, including electronegativity and ionization energy. A detailed correlation analysis between prompt dimensions and elemental properties is provided in Appendix~\ref{expressiveness_node}.\\
For additional analysis, we visualize the node-level prompt embeddings using t-SNE~\cite{van2008visualizing} (Fig.~\ref{fig:node_prompt_interpret}), coloring elements by their periodic block. The resulting embedding space exhibits clear structure: f-block elements form a compact, well-separated cluster; s-block elements appear as distinct outliers; and d- and p-block elements occupy broader, continuous regions with partial overlap. This organization aligns with known chemical principles, where certain element groups are sharply distinguished while others exhibit more gradual transitions. To quantify this structure, we perform k-means clustering on the prompt embeddings, obtaining a clustering purity of 0.73 and a Normalized Mutual Information (NMI) of 0.64 against ground-truth periodic-block labels. Together, these qualitative and quantitative results provide complementary evidence that the learned prompt embeddings capture chemically meaningful relationships beyond simple correlation.

\begin{table}[t]
\renewcommand\arraystretch{1.}
\centering
\resizebox{0.90\columnwidth}{!}{
\begin{tabular}{ccccccp{0.5cm}<{\centering}p{0.9cm}<{\centering}p{0.9cm}<{\centering}p{0.9cm}<{\centering}p{0.9cm}<{\centering}p{0.9cm}<{\centering}p{0.9cm}<{\centering}p{0.9cm}<{\centering}cc} \toprule
\makecell{Encoder \\ Strategy} & \makecell{Tuning \\ Strategy} & \makecell{Formation \\ Energy} & \makecell{Bandgap \\ (OPT)} & \makecell{Total \\ Energy} & \makecell{Ehull\\{}} \\ \midrule

\multirow{4}{*}{CGCNN} &  Vanilla & 0.063 & 0.200 & 0.078 & 0.170\\
~ & Only Node & \textbf{0.041} & 0.160 & \ul{0.048} & \ul{0.117}\\
~ & Only Graph & \ul{0.042} & \ul{0.156} & 0.049 & 0.119 \\ 

~ & Both & \textbf{0.041} & \textbf{0.155} & \textbf{0.047} & \textbf{0.116}\\

\midrule

\multirow{4}{*}{ALIGNN} &  Vanilla & 0.035 & 0.142 & 0.037 & 0.076\\
~ & Only Node & \ul{0.034} & \ul{0.141} & \textbf{0.035} & 0.066 \\
~ & Only Graph & \ul{0.034} & \ul{0.141} & 0.036 & \ul{0.065} \\ 

~ & Both & \textbf{0.033} & \textbf{0.140} & \textbf{0.035} & \textbf{0.064}\\

\midrule

\multirow{4}*{PotNet} & Vanilla &\ul{0.0294} & \textbf{0.127} & 0.0321 & 0.055\\

~ & Only Node & \ul{0.0294} & \textbf{0.127} & 0.0317 & 0.053\\
~ & Only Graph & 0.0299 & \ul{0.129} & \ul{0.0314} & \ul{0.049}\\

~ & Both & \textbf{0.0292} & \textbf{0.127} & \textbf{0.0310} & \textbf{0.048}\\
\bottomrule
\end{tabular}
}
\caption{Ablation results with different modalities of prompt learning strategies. Best results is \textbf{bold} and the second best result is \ul{underlined}.}
\label{tab:ablation_1_new}
\end{table}
\begin{figure*}[!ht]%
    \centering
    {\includegraphics[width=1.0\textwidth]{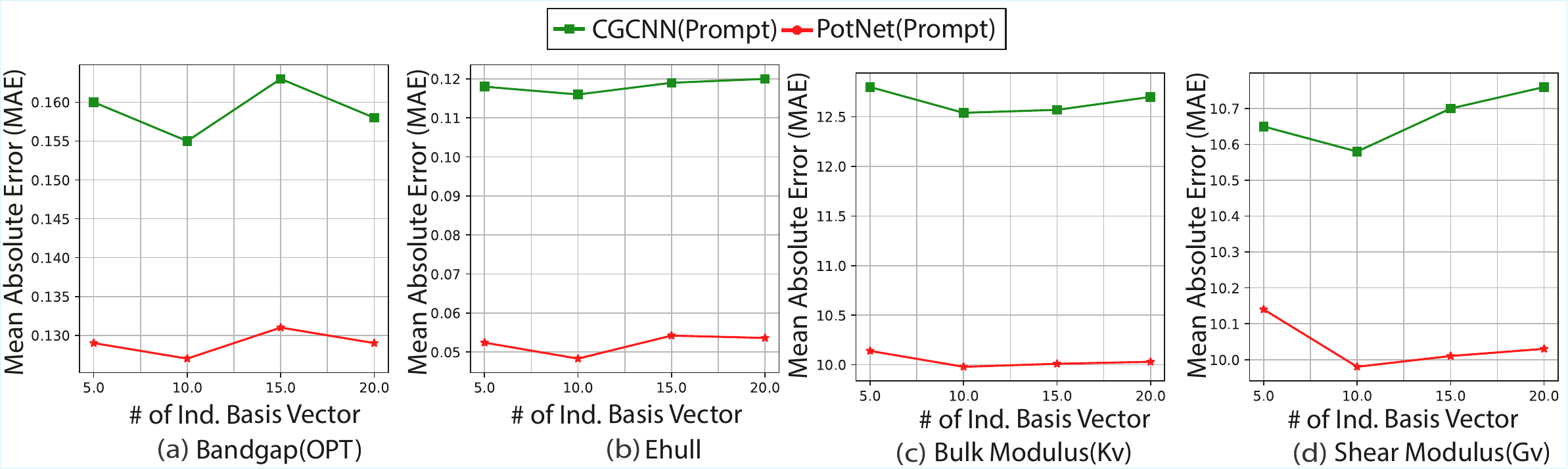}}
    {\caption{This illustration shows the sensitivity analysis on the choice of independent basis vectors on four properties (a) Bandgap(OPT), (b) Ehull, (c) Bulk Modulus(Kv), and (d) Shear Modulus(Gv) using JARVIS-DFT dataset}
    \label{fig:senstivity_ablation}}
\end{figure*}
\subsection{Impact of Different Prompt Modalities}
\label{ablation_prompt_modality}
We conduct an ablation study to investigate the influence of different prompt modalities in enhancing SOTA model performance and report the results in Table~\ref{tab:ablation_1_new}. Specifically, we conduct experiments with three setups: 1) By default, in our proposed framework we infused both node and graph-level prompts (denoted as \emph{``Both"}), 2) With only node-level prompts infused at input node-features and no graph-level prompts (denoted as \emph{``Only Node"}), 3) With only graph-level prompts infused at graph embedding and no node-level prompts (denoted as \emph{``Only Graph"}). For this study, we use three baseline models: CGCNN, ALIGNN, and PotNet, each trained on four properties from the JARVIS-DFT dataset: formation energy, bandgap (OPT), total energy, and Ehull. It is evident that using both node-level and graph-level prompt learning together (Both), CGCNN, ALIGNN, and PotNet yields the best results in most cases, surpassing the performance of using node-level or graph-level prompt learning alone. This indicates that integrating both node-level and graph-level prompts effectively captures atomic chemical semantics and crystal topological information, leading to enhanced property prediction performance. Additional results for other properties can be found in the Appendix \ref{appendix_prompt_modality}.
\subsection{Sensitivity Analysis on No. Basis Vectors.}
\label{appendix_basis}
We also conducted a sensitivity analysis on \(k\) (the number of independent basis vectors) used in node-prompt learning (Section ~\ref{sec_node_prompt}). For this, we performed an ablation study using the prompt fused variant of CGCNN and PotNet on four properties from the JARVIS-DFT dataset. Experiments were conducted with varying values of \(k\), and the results are presented in Fig.~\ref{fig:senstivity_ablation}. From the figure, we observe that the best MAE for all properties is achieved with $k = 10$. However, increasing \(k\) to 15 or 20 does not yield optimal results. Consequently, we selected 10 independent basis vectors for node-level prompts in all our experiments, as presented before.\\\\
\xhdr{Additional Results} Appendix (\ref{appendix_results}) provides details on additional experiments, including a alternative architectural modifications (\ref{arch_modification_appendix}), impact of different prompt modalities (\ref{appendix_prompt_modality}), statistical significance of the results (\ref{statistical-test}), efficiency analysis (\ref{time_complex}), robustness to label noise~\ref{robust_ablation}, contribution of graph-level prompt~\ref{graph-prompt}, fairness of baseline prompt methods~\ref{fairness_comparison}, expressiveness of node-level prompt~\ref{expressiveness_node} and expressiveness of graph-level prompt (\ref{expressiveness_graph}).

\section{Conclusion}
Crystal property prediction remains a challenging task in materials science and recent research focused on encoding intrinsic domain knowledge into GNN encoders by developing increasingly complex architectures. However, this approach often leads to a significant increase in parameters, resulting in computational overheads. In this work, we introduce soft prompt learning, a novel paradigm designed to capture latent atomic and structural features in GNN models, thereby enhancing crystal property prediction performance but with minimal increase in parameter space. Our method incorporates learnable prompts at both the node and graph levels. At the node level, soft prompts are tailored to capture the local chemical semantics of constituent atoms, while at the graph level, they encode the underlying crystal symmetry. Through extensive experiments on benchmark datasets, we demonstrate that our framework significantly improves performance across various SOTA GNN models. Notably, we observe that incorporating soft prompts enhances the expressive power of both simpler models like CGCNN and advanced models such as ALIGNN, Matformer, and PotNet. Furthermore, when compared to alternative architectural modification strategies, prompt learning proves to be both more effective and lightweight. Additionally, we show that soft prompts possess cross-property knowledge transfer capabilities, enabling improved performance in limited data scenarios by transferring learned knowledge across different properties. We believe that this soft prompt learning paradigm for GNNs has the potential to extend beyond the crystal domain, enhancing the performance of state-of-the-art GNN models in a wide range of downstream tasks.

\bibliography{uai}

\newpage

\onecolumn

\title{Model Agnostic Graph Prompt Learning for Crystal Property Prediction\\(Supplementary Material)}
\maketitle

\appendix
\section{Multi-graph Construction for Crystals}
\label{crystal_multigraph}
\begin{sloppypar}
We realize a crystal material as a multi-graph structure $\mathcal{G} =(\mathcal{V}, \mathcal{E}, \mathcal{X}, \mathcal{F} )$ as shown in Fig.~\ref{fig:periodic_crystals} (c)~\cite{xie2018crystal}. $\mathcal{G}$ is an undirected weighted multi-graph where $\mathcal{V}$ denotes the set of nodes or atoms present in a unit cell of the crystal structure. $\mathcal{E}=\{(u,v,k_{uv})\}$ denotes a multi-set of node pairs and $k_{uv}$ denotes number of edges between a node pair $(u,v)$. $\mathcal{X}=\{(x_{u} | u \in \mathcal{V} )\}$ denotes 92 dimensional node feature set proposed by CGCNN~\cite{xie2018crystal}. It includes different chemical properties like electronegativity, valence electron, covalent radius, etc. Finally, $\mathcal{F}=\{\{s^k\}_{(u,v)} |  (u,v) \in \mathcal{E}, k\in\{1..k_{uv}\}\}$ denotes the multi-set of edge weights where $s^k$ corresponds to the $k^{th}$ bond length between a node pair $(u,v)$, which signifies the inter-atomic bond distance between two atoms.
\end{sloppypar}
\section{GNN for Crystal Materials}
\label{gnn_crystal}
Graph neural network models have been very effective in learning crystal representation. CGCNN \cite{xie2018crystal} is the first proposed model, which represents 3D crystal structure as an undirected weighted multi-edge graph and applies a graph convolution neural network to update node features based on their local chemical and structural environment. The fundamental principle of a GNN model is the message passing and aggregation paradigm, where it learns node representations by recursively aggregating the representations (messages) from neighboring nodes within the graph. Formally, at $l^{th}$ layer of a GNN model, node representation of node $u$ is updated as follows :
\begin{equation}
    h^{(l+1)}_u = COMB \biggl\{ h^{(l)}_u, \underset{v \in N_u}{AGG}\biggl(h^{(l)}_v,h^{(l)}_u,{e^{(l)}_{(u,v)}}_k \biggr) \biggr\}
\end{equation}
where $h^{(l)}_u,h^{(l)}_v$ denote node representation of nodes $u$ and $v$, respectively, at $l^{th}$ layer, ${e^{(l)}_{(u,v)}}_k$ denotes representation of $k^{th}$ edge between $(u,v)$ at $l^{th}$ layer and $N_u$ denotes the set of neighboring nodes of node $u$. The AGG function aggregates messages from the neighboring nodes of $u$, while the COMB function combines this aggregated message with the node representation of $u$ to produce the node representation of $u$ for the next $(l+1)^{th}$ layer. After applying $L$ such layers of aggregation and combining, the final set of node embeddings $\mathcal{H} = \{h_1,..., h_{|\mathcal{V}|}\}$ is produced, where $h_u:=h_u^{(L)} \in \mathbb{R}^{d}$ represents the final embedding of node $u$.

\section{Related Work}
\subsection{Crystal Property Prediction}
Historically, Density Functional Theory (DFT)~\cite{orio2009density} has been an effective tool for both crystal material generation and estimating several properties of crystals by examining the electronic density of their atomic configurations. However, DFT calculation requires substantial computation costs; hence, it poses a significant challenge for novel materials discovery in the vast materials space. Recently, thanks to the advances of deep learning, data-driven~\cite{seko2015prediction,pilania2015structure,lee2016prediction,de2016statistical,seko2017representation,isayev2017universal,ward2017including,lu2018accelerated,im2019identifying} techniques have become quite popular to establish relationships between crystal atomic structure and their properties effectively. 

Particularly, for material generation, equivariant diffusion models~\cite{xie2021crystal,luo2023towards,jiao2023crystal,yang2023scalable,jiao2024space,miller2024flowmm}, Flow Models~\cite{lipman2023flow, albergo2023building, liu2023flow,das2025periodic} and Large Language Model~\cite{gruver2024fine, khastagir2026llm,govindarajan2026crystune} been explored. On the contrary, graph neural network (GNN)~\cite{xie2018crystal,chen2019graph,louis2020graph,Wolverton2020,schmidt2021crystal,choudhary2021atomistic,yan2022periodic,lin2023efficient} based approaches become popular for crystal property prediction to encode complex crystal structures along with atom features in an enriched representation space. 

Earlier approaches~\cite{xie2018crystal, chen2019graph, louis2020graph, Wolverton2020, schmidt2021crystal} construct a multi-edge graph from the 3D crystal structure and apply a GNN model to encode the neighborhood structural information around an atom. Building on this, numerous studies have proposed various GNN architecture variants that integrate domain-specific knowledge into the encoder to improve crystal representation learning. ALIGNN~\cite{choudhary2021atomistic} incorporates bond angular information among edges to capture many-body interactions; Matformer~\cite{yan2022periodic} is designed to be invariant to periodicity, enabling it to explicitly capture repeating patterns; and PotNet~\cite{lin2023efficient} models inter-atomic potentials directly based on physical principles. In a related direction, CrysMMNet~\citep{das2023crysmmnet} predicts material properties by fusing textual descriptions with graph-based structural representations. Moreover, data scarcity is a known challenge in this field, and to address it, several graph pre-training strategies have been introduced. CrysXPP~\cite{das2022crysxpp} is the first pre-trained/fine-tune model, which uses an autoencoder to train on a volume of un-tagged crystal graphs (unsupervised pre-training) and the learned knowledge is used to initialize the encoder of a property predictor, which is further fine-tuned with property-specific tagged data. Following this work, CrysGNN~\cite{das2023crysgnn} proposed large-scale pre-training with a huge corpus of curated unlabelled data and incorporates the idea of knowledge distillation to distill the important information from the pre-trained model and injects it into the property prediction process. Building on this line of work, Crystal Twins~\citep{magar2022crystal} takes a self-supervised route, using a Barlow Twins-style objective to pre-train a CGCNN backbone. A separate thread of research turns to diffusion models for learning crystal representations through structure reconstruction. CrysDiff~\citep{song2024diffusion}, for instance, pre-trains a joint denoising diffusion model that reconstructs crystal geometry starting only from atomic composition; at finetuning time, the structure is held fixed and the diffusion process is instead conditioned on the target property. DPF~\citep{shen2025denoising} follows a related but distinct recipe: it corrupts atom identities, atomic positions, and lattice parameters during pre-training and trains the model to recover the original structure, after which the resulting representations are finetuned for property-prediction tasks. CrysAtom~\citep{mukherjee2025crysatom} takes yet another approach, learning distributed atom-level embeddings from unlabeled crystals in an unsupervised manner and shows this pretraining strategy meaningfully improves downstream property-prediction performance. Most recently, CrysLDNet~\citep{mukherjee2026latent} departs from operating directly in the raw feature space and instead moves pretraining into a compact latent space: a variational autoencoder first compresses the crystal structure into a smooth latent representation, and a diffusion model is then trained to denoise within that space rather than over atom types, coordinates, and lattice parameters directly. The pretrained encoder is subsequently finetuned with a lightweight prediction head for downstream property prediction. 
\subsection{Graph-Prompt Tuning Methods}
Prompt tuning methods, originating from Natural Language Processing (NLP), have been widely used to facilitate the adaptation of pre-trained language models to various downstream tasks~\cite{liu2021pre}. Particularly, research has shown using soft prompts model can achieve optimal performance in various downstream tasks~\cite{liu2021pre,lester2021power}. Since, there are widely available pre-trained models like BERT~\cite{devlin2018bert}, GPT~\cite{radford2019language}, etc in NLP, traditional prompt-based approaches~\cite{liu2021pre,lester2021power} freeze the parameters of the pre-train model and introduce additional learnable components in the input space, thereby enhancing the compatibility between inputs and pre-trained models. Further, in computer vision, Jia et al.~\cite{jia2022visual} and Bahng et. al.~\cite{bahng2022exploring} investigate the efficacy of adapting large-scale methods by modifying input images at the pixel level. \\
Inspired by this, few recent studies have explored prompt-tuning methods for graph neural networks(GNNs). GPPT~\cite{sun2022gppt} proposes the learnable edge-level prompts, transforming node classification into a link prediction task to mitigate the structural gap. Graphprompt~\cite{liu2023graphprompt} is developed to unify graph prompt templates and enhance their performance through learnable readout prompt functions. All-in-One~\cite{sun2023all} proposed a novel graph token structure accompanied to aligning pre-training and fine-tuning objectives in the context of graph-based tasks. GPF~\cite{fang2024universal} integrates the feature prompt in a more general way to improve the prompt performance. HGPrompt~\cite{yu2024hgprompt} designs meta-path-based prompts for heterogeneous graphs. SUPT~\cite{lee2024subgraph} proposes subgraph-level universal graph prompt learning, while LEAP~\cite{xu2025learning} introduces a reinforcement learning (RL) based approach for editing universal graph prompts in molecular graphs. However, leveraging prompt-based techniques for crystal property prediction remains largely unexplored by the research community. Also, current graph prompt-tuning methods operate within the pre-train/fine-tune paradigm, and to the best of our knowledge, no prior work has explored prompt-tuning for supervised learning from scratch.

\section{Additional Experimental Results}
\label{appendix_results}
\begin{table*}[!ht]
\renewcommand\arraystretch{0.9}
\setlength{\tabcolsep}{2.5pt}
\centering
\resizebox{\textwidth}{!}{
\begin{tabular}{clccccccccccp{0.1cm}<{\centering}p{0.1cm}<{\centering}p{0.1cm}<{\centering}p{0.1cm}<{\centering}p{0.1cm}<{\centering}p{0.1cm}<{\centering}p{0.1cm}<{\centering}p{0.1cm}<{\centering}cc<{\centering}cc} \toprule
\makecell{GNN \\ Model} & \makecell{Setup \\ {}} & \makecell{\# Parameters \\ {}} & \makecell{Formation \\ Energy} & \makecell{Bandgap \\ (OPT)} & \makecell{Total \\ Energy} & \makecell{Ehull\\{}} & \makecell{Bandgap \\ (MBJ)} & \makecell{Bulk Modulus \\ (Kv)} & \makecell{Shear Modulus\\ (Gv)} & \makecell{SLME \\ (\%)} & \makecell{Spillage \\ {}}  
\\ 
\midrule

\multirow{6}{*}{} &  Vanilla & 0.423M & 0.063 & 0.200 & 0.078 & 0.170 & 0.410 & 14.47 & 11.75 & 5.66 & 0.397\\
~ & 2 $\times$ GNN Layers & 0.788M & 0.061 & 0.174 & 0.073 & 0.162 & 0.429 & 14.15 & 11.39 & 5.60 & 0.402\\

~ &  Increase Hidden Dim & 1.435M & 0.055 & 0.177 & 0.072 & 0.175 & 0.419 & 14.74 & 11.59 & 5.31 & 0.416\\

CGCNN & Skip Connection & 0.423M & 0.059 & 0.192 & 0.073 & 0.161 & 0.404 & 15.37 & 12.30 & 5.66 & 0.429\\

~ & Adapter Learning & 0.476M & 0.151 & 0.224 & 0.278 & 0.261 & 0.512 & 14.62 & 11.36 & 4.98 & 0.410\\

~ & Prompt Learning & 0.433M & \textbf{0.041} & \textbf{0.155} & \textbf{0.047} & \textbf{0.116} & \textbf{0.354} & \textbf{12.54} & \textbf{10.58} & \textbf{4.81} & \textbf{0.388}\\

\midrule

\multirow{5}{*}{} & Vanilla & 3.27M & 0.033 & 0.137 & 0.035 & 0.064 & 0.300 & 11.21 & 10.76 & 5.26 & 0.398\\
~ & 2 $\times$ GNN Layers & 6.34M & 0.032 & 0.143 & 0.035 & 0.057 & 0.316 & 10.42 & 9.60 & 5.11 & 0.367\\

~ &  Increase Hidden Dim & 12.84M & 0.052 & 0.139 & 0.092 & 0.104 & 0.328 & 10.85 & 9.93 & 4.76 & 0.367\\

Matformer & Skip Connection & 3.27M & 0.033 & 0.138 & 0.034 & 0.064 & 0.313 & 10.81 & 10.78 & 4.87 & 0.368\\

~ & Adapter Learning & 3.36M & 0.089 & 0.156 & 0.194 & 0.155 & 0.360 & 13.28 & 11.02 & 5.15 & 0.380\\

~ & Prompt Learning & 3.28M & \textbf{0.032} & \textbf{0.136} & \textbf{0.033} & \textbf{0.053} & \textbf{0.285} & \textbf{10.05} & \textbf{9.57} & \textbf{4.76} & \textbf{0.360}\\

\bottomrule
\end{tabular}
}
\caption{Summary of results (MAE) for various properties in JARVIS-DFT using different architectural modifications in CGCNN and Matformer. Our proposed prompt learning paradigm is both lightweight and delivers superior performance compared to alternative approaches.}
\label{tab:Different_modalities_appendix}
\end{table*}
\begin{table*}[!ht]
\renewcommand\arraystretch{0.9}
\centering
\resizebox{0.95\textwidth}{!}{
\begin{tabular}{cccccccccccp{0.5cm}<{\centering}p{0.9cm}<{\centering}p{0.9cm}<{\centering}p{0.9cm}<{\centering}p{0.9cm}<{\centering}p{0.9cm}<{\centering}p{0.9cm}<{\centering}p{0.9cm}<{\centering}cc} \toprule
\makecell{Encoder \\ Strategy} & \makecell{Tuning \\ Strategy} & \makecell{Formation \\ Energy} & \makecell{Bandgap \\ (OPT)} & \makecell{Total \\ Energy} & \makecell{Ehull\\{}} & \makecell{Bandgap \\ (MBJ)} & \makecell{Bulk \\ Modulus(Kv)} & \makecell{Shear \\ Modulus(Gv)} & \makecell{SLME \\ (\%)} & \makecell{Spillage \\ {}}  
\\ 
\midrule

\multirow{4}{*}{CGCNN} &  Vanilla & 0.063 & 0.200 & 0.078 & 0.170 & 0.410 & 14.47 & 11.75 & 5.66 & 0.397\\
~ & Only Node & \textbf{0.041} & 0.160 & \ul{0.048} & \ul{0.117} & \ul{0.344} & \ul{12.35} & \textbf{10.38} & \textbf{4.76} & \textbf{0.384}\\
~ & Only Graph & \ul{0.042} & \ul{0.156} & 0.049 & 0.119 & \textbf{0.341} & \textbf{12.25} & \ul{10.51} & 4.92 & 0.391\\ 
~ & Both & \textbf{0.041} & \textbf{0.155} & \textbf{0.047} & \textbf{0.116} & 0.354 & 12.54 & 10.58 & \ul{4.80} & \ul{0.388}\\

\midrule

\multirow{4}{*}{ALIGNN} &  Vanilla & 0.035 & 0.142 & 0.037 & 0.076 & 0.310 & 10.40 & 9.86 & 4.52 & 0.351\\
~ & Only Node & \ul{0.034} & \ul{0.141} & \textbf{0.035} & 0.066 & \ul{0.292} & \ul{10.24} & \ul{9.71} & 4.49 & \textbf{0.346}\\
~ & Only Graph & \ul{0.034} & \ul{0.141} & \ul{0.036} & \ul{0.065} & \textbf{0.291} & 10.38 & \textbf{9.62} & \ul{4.47} & 0.350 
\\ 

~ & Both & \textbf{0.033} & \textbf{0.140} & \textbf{0.035} & \textbf{0.064} & 0.299 & \textbf{10.21} & 9.73 & \textbf{4.42} & \ul{0.348}\\

\midrule

\multirow{4}{*}{Matformer} &  Vanilla & \ul{0.033} & \ul{0.137} & 0.035 & 0.064 & 0.300 & 11.21 & 10.76 & 5.26 & 0.398\\
~ & Only Node & \textbf{0.032} & \textbf{0.136} & \ul{0.034} & \ul{0.056} & \ul{0.295} & \ul{10.21} & \textbf{9.53} & 4.77 & \ul{0.359}\\
~ & Only Graph & \ul{0.033} & 0.138 & \ul{0.034} &  \ul{0.056} & 0.297 & 10.26 & 9.62 & \textbf{4.65} & \textbf{0.355}\\ 
~ & Both & \textbf{0.032} & \textbf{0.136} & \textbf{0.033} & \textbf{0.053} & \textbf{0.285} & \textbf{10.05} & \ul{9.57} & \ul{4.76} & 0.360\\

\midrule

\multirow{4}*{PotNet} & Vanilla &\ul{0.0294} & \textbf{0.127} & 0.0321 & 0.055 & 0.270 & 10.11 & 9.23 & 4.57 & 0.361\\

~ & Only Node & \ul{0.0294} & \textbf{0.127} & 0.0317 & 0.053 & \textbf{0.251} & \ul{9.95} & \textbf{8.80} & \ul{4.38} & \textbf{0.349}\\
~ & Only Graph & 0.0299 & \ul{0.129} & \ul{0.0314} & \ul{0.049} & 0.266 & \textbf{9.64} & \ul{8.98} & \textbf{4.22} & \ul{0.350}\\

~ & Both & \textbf{0.0292} & \textbf{0.127} & \textbf{0.0310} & \textbf{0.048} & \ul{0.258} & 9.98 & 9.01 & 4.43 & 0.352\\
\bottomrule
\end{tabular}
}
\caption{{Ablation results with different modalities of prompt learning strategies. Best results is \textbf{bold} and the second best result is \ul{underlined}.}}
\label{tab:ablation_1}
\end{table*}
\subsection{Alternate Architectural Modifications}
\label{arch_modification_appendix}
Next, we explore alternative methods for learning latent chemical and structural features by making simple modifications to the existing GNN models. The approaches investigated are increasing hidden layer dimensions, doubling the number of GNN layers, introducing skip connections, and implementing adapter learning. 
\begin{enumerate}[nolistsep]
    \setlength\itemsep{0.3em}
    \item \textbf{2 $\times$ GNN Layers.} In this variant, the number of GNN layers is increased by doubling the original number of layers.
    \item \textbf{Increasing Hidden Dim.} In this variant, the hidden layer dimensions for each GNN are increased by 64.
    \item \textbf{Skip Connection.} In this variant, residual connections are added to each layer of the GNN.
    \item \textbf{Adapter Learning.} In this variant, a parameter-efficient adapter-based tuning technique is used for each layer of the GNN. The method follows the approach proposed in AdapterGNN~\cite{li2024adaptergnn}, where parameter-efficient adapters are added to each layer of the GNN models.
\end{enumerate}
We conduct another set of experiments using JARVIS-DFT dataset on CGCNN and Matformer to compare the performance of these alternative methods with prompt learning and report the results in Table~\ref{tab:Different_modalities_appendix}. We observe that compared to all other approaches, prompt learning introduces minimal parameter overhead while outperforming all of these. In particular, increasing hidden layer dimensions or doubling the number of GNN layers substantially increases the model's parameter count, yet results in minimal performance improvement across all properties. On the other hand, while adapter learning and skip connections add only a small parameter overhead, they also fail to significantly enhance crystal property prediction performance.
This highlights the ingenuity of our graph prompt learning approach, which is lightweight, model-agnostic, and provides substantial improvement over the vanilla models.

\subsection{Impact of Different Prompt Modalities.}
\label{appendix_prompt_modality}
In the main paper, a brief summary of this result is provided; here, we present an elaboration on some of the details. We conduct an ablation study to investigate the influence of different prompt modalities in enhancing SOTA model performance and report the results in Table~\ref{tab:ablation_1}. In specific, we conduct experiments with three setups. 1) By default, in our proposed framework we infused both node and graph-level prompts (denoted as \emph{"Both"}) 2) With only node-level prompts infused at input node-features and no graph-level prompts (denoted as \emph{"Only Node"}) 3) With only graph-level prompts infused at graph embedding and no node-level prompts (denoted as \emph{"Only Graph"}) For this study, we used four baseline models: CGCNN, ALIGNN, Matformer and PotNet, each trained on nine properties from the JARVIS-DFT dataset: formation energy, bandgap (OPT), total energy, Ehull, bandgap(MBJ), bulk modulus(Kv), shear modulus(Gv), SLME(\%), and Spillage. From Table~\ref{tab:ablation_1} it is evident that using both node-level and graph-level prompt learning together (Both), CGCNN, ALIGNN, Matformer, and PotNet yields the best results in most of the cases (15 out of 36 experiments), surpassing the performance of using either node-level or graph-level prompt learning alone.  
\begin{table*}[!htbp]
\centering
\setlength\tabcolsep{3pt}
\renewcommand{\arraystretch}{1.05}

\subfloat[Comparison for CGCNN and ALIGNN backbones.]{
\resizebox{0.95\textwidth}{!}{
\begin{tabular}{l|cccc|cccc}
\toprule
{Property} & {CGCNN} & \multicolumn{3}{c|}{\shortstack{CGCNN\\(Prompt)}} & {ALIGNN} & \multicolumn{3}{c}{\shortstack{ALIGNN\\(Prompt)}} \\
\cmidrule(lr){3-5} \cmidrule(lr){7-9}
 &  & Mean$\pm$Std & CI & P-Value &  & Mean$\pm$Std & CI & P-Value \\
\midrule
Formation Energy & 0.063 & {0.041$\pm$0.010} & (0.0286, 0.0534) & 0.0079 & 0.035 & {0.033$\pm$0.001} & (0.0318, 0.0342) & 0.0110 \\

Bandgap & 0.200 & {0.155$\pm$0.020} & (0.1302, 0.1798) & 0.0073 & 0.142 & {0.140$\pm$0.001} & (0.1388, 0.1412) & 0.0110 \\

Total Energy & 0.078 & {0.047$\pm$0.010} & (0.0346, 0.0594) & 0.0023 & 0.037 & {0.035$\pm$0.0009} & (0.0339, 0.0361) & 0.0077 \\

Ehull & 0.170 & {0.116$\pm$0.020} & (0.0912, 0.1408) & 0.0038 & 0.076 & {0.064$\pm$0.004} & (0.0590, 0.0690) & 0.0026 \\

mbj Bandgap & 0.410 & {0.354$\pm$0.010} & (0.3416, 0.3664) & 0.0002 & 0.310 & {0.299$\pm$0.003} & (0.2953, 0.3027) & 0.0012 \\

Bulk Modulus & 14.470 & {12.540$\pm$1.100} & (11.1742, 13.9058) & 0.0172 & 10.400 & {10.210$\pm$0.110} & (10.0734, 10.3466) & 0.0181 \\

Shear Modulus & 11.750 & {10.580$\pm$0.300} & (10.2075, 10.9525) & 0.0010 & 9.862 & {9.731$\pm$0.090} & (9.6193, 9.8427) & 0.0312 \\

SLME & 5.660 & {4.801$\pm$0.400} & (4.3043, 5.2977) & 0.0086 & 4.521 & {4.420$\pm$0.060} & (4.3455, 4.4945) & 0.0197 \\

Spillage & 0.397 & {0.388$\pm$0.005} & (0.3818, 0.3942) & 0.0158 & 0.351 & {0.348$\pm$0.002} & (0.3455, 0.3505) & 0.0285 \\
\bottomrule
\end{tabular}}
}
\\
\subfloat[Comparison for Matformer and PotNet backbones.]{
\resizebox{0.95\textwidth}{!}{
\begin{tabular}{l|cccc|cccc}
\toprule
{Property} & {Matformer} & \multicolumn{3}{c|}{\shortstack{Matformer\\(Prompt)}} & {PotNet} & \multicolumn{3}{c}{\shortstack{PotNet\\(Prompt)}} \\
\cmidrule(lr){3-5} \cmidrule(lr){7-9}
 &  & Mean$\pm$Std & CI & P-Value &  & Mean$\pm$Std & CI & P-Value \\
\midrule
Formation Energy & 0.033 & 0.032$\pm$0.0007 & (0.0311, 0.0329) & 0.033 & 0.0294 & 0.0292$\pm$0.0001 & (0.0291, 0.0293) & 0.011 \\

Bandgap & 0.137 & 0.136$\pm$0.0005 & (0.1354, 0.1366) & 0.011 & 0.1274 & 0.1270$\pm$0.0002 & (0.1268, 0.1272) & 0.011 \\

Total Energy & 0.035 & 0.033$\pm$0.001 & (0.0318, 0.0342) & 0.011 & 0.0324 & 0.0311$\pm$0.001 & (0.0299, 0.0323) & 0.044 \\

Ehull & 0.064 & 0.053$\pm$0.004 & (0.0480, 0.0580) & 0.0035 & 0.0552 & 0.0481$\pm$0.004 & (0.0431, 0.0531) & 0.0166 \\

mbj Bandgap & 0.300 & 0.285$\pm$0.010 & (0.2726, 0.2974) & 0.028 & 0.2701 & 0.258$\pm$0.007 & (0.2493, 0.2667) & 0.018 \\

Bulk Modulus & 11.21 & 10.05$\pm$0.720 & (9.1560, 10.9440) & 0.0227 & 10.11 & 9.981$\pm$0.090 & (9.8693, 10.0927) & 0.0327 \\

Shear Modulus & 10.76 & 9.581$\pm$0.511 & (8.9465, 10.2155) & 0.0067 & 9.232 & 9.0201$\pm$0.102 & (8.8935, 9.1467) & 0.0097 \\

SLME & 5.26 & 4.764$\pm$0.322 & (4.3642, 5.163) & 0.026 & 4.5701 & 4.431$\pm$0.104 & (4.3019, 4.5601) & 0.040 \\

Spillage & 0.398 & 0.359$\pm$0.020 & (0.3342, 0.3838) & 0.012 & 0.361 & 0.3522$\pm$0.003 & (0.3485, 0.3559) & 0.0028 \\
\bottomrule
\end{tabular}}
}
\\
\subfloat[Comparison for iComFormer and eComFormer backbones.]{
\resizebox{0.95\textwidth}{!}{
\begin{tabular}{l|cccc|cccc}
\toprule
{Property} & {iComFormer} & \multicolumn{3}{c|}{\shortstack{iComFormer\\(Prompt)}} & {eComFormer} & \multicolumn{3}{c}{\shortstack{eComFormer\\(Prompt)}} \\
\cmidrule(lr){3-5} \cmidrule(lr){7-9}
 &  & Mean$\pm$Std & CI & P-Value &  & Mean$\pm$Std & CI & P-Value \\
\midrule
Formation Energy & 0.027 & 0.026$\pm$0.0003 & (0.0256, 0.0264) & 0.0017 & 0.028 & 0.027$\pm$0.0003 & (0.0266, 0.0274) & 0.0017 \\

Bandgap & 0.135 & 0.134$\pm$0.0004 & (0.1335, 0.1345) & 0.0050 & 0.124 & 0.122$\pm$0.001 & (0.1208, 0.1232) & 0.0111 \\

Total Energy & 0.031 & 0.029$\pm$0.001 & (0.0279, 0.0301) & 0.0077 & 0.032 & 0.031$\pm$0.0002 & (0.0308, 0.0312) & 0.00036 \\

Ehull & 0.047 & 0.046$\pm$0.0002 & (0.0458, 0.0462) & 0.00036 & 0.044 & 0.040$\pm$0.001 & (0.0388, 0.0412) & 0.00086 \\

mbj Bandgap & 0.264 & 0.259$\pm$0.003 & (0.2553, 0.2627) & 0.0204 & 0.280 & 0.276$\pm$0.002 & (0.2735, 0.2785) & 0.0111 \\

Bulk Modulus & 14.27 & 13.22$\pm$0.72 & (12.3260, 14.1140) & 0.0311 & 13.45 & 12.26$\pm$0.825 & (11.2356, 13.2844) & 0.0321 \\

Shear Modulus & 9.76 & 9.359$\pm$0.201 & (9.1094, 9.6086) & 0.0112 & 11.12 & 10.56$\pm$0.32 & (10.1627, 10.9573) & 0.0173 \\

SLME & 4.702 & 4.342$\pm$0.23 & (4.0564, 4.6276) & 0.0249 & 4.662 & 4.650$\pm$0.004 & (4.6450, 4.6550) & 0.0026 \\

Spillage & 0.369 & 0.351$\pm$0.01 & (0.3386, 0.3634) & 0.0158 & 0.382 & 0.369$\pm$0.009 & (0.3578, 0.3802) & 0.0320 \\
\bottomrule
\end{tabular}}
}
\caption{Statistical comparison of \textit{prompt} version across six backbone models. }
\label{tab:crysldnet_subtables}
\end{table*}

\subsection{Statistical Significance of The Results}
\label{statistical-test}
We perform a comprehensive statistical analysis to ensure the robustness and reliability of the reported performance improvements. For each prompt-enhanced variant, we conduct five independent runs with different random seeds and report the mean, standard deviation, and 95\% confidence interval (CI). In addition, we compute paired t-test p-values for each backbone to assess the statistical significance of the observed improvements. Specifically, we evaluate CGCNN, ALIGNN, Matformer, PotNet, iComFormer, and eComFormer, along with their corresponding prompt-enhanced variants that use these encoders as backbones. The complete results on the JARVIS dataset are presented in Table~\ref{tab:crysldnet_subtables}.
\\
These statistical measures demonstrate that the performance gains achieved by the prompt-enhanced models are consistent and reproducible across multiple runs. Notably, the paired t-tests yield p-values below 0.05 for most evaluated properties, indicating statistically significant improvements. Overall, this analysis confirms that the improvements obtained through prompt learning do not arise from random variation but instead reflect genuine and meaningful performance gains across downstream tasks.
\begin{table}[t]
\centering
\resizebox{0.85\columnwidth}{!}{
\begin{tabular}{llcccc}
\toprule
Backbone & Method & Params (M) & Total Training & Time/Epoch (s) & Total Training\\
& & & FLOPs(G) &  & Time(s)\\ 
\midrule

\multirow{4}{*}{CGCNN}
& GPF & 0.323M & 2.53$\times10^6$ & 14.63 & 14,630 \\
& SUPT & 0.326M & 2.90$\times10^6$ & 29.33 & 29,334 \\
& LEAP & 0.954M & 1.65$\times10^8$ & 1250 & 1,250,000 \\
& Prompt (Ours) & 0.433M & 2.54$\times10^6$ & 14.74 & 14,740 \\

\midrule

\multirow{4}{*}{Matformer}
& GPF & 2.18M & 2.10$\times10^{6}$ & 85 & 85,000\\
& SUPT & 11.5M & 2.80$\times10^{6}$ & 110 & 110,000\\
& LEAP & 19.4M & 1.80$\times10^{8}$ & 4043.4 & 4,043,400\\
& Prompt (Ours) & 3.28M & 2.27$\times10^{6}$ & 90 & 90,000 \\

\bottomrule
\end{tabular}}
\caption{Efficiency comparison of graph prompt learning methods on crystal property prediction tasks.}
\label{tab:efficiency}
\end{table}
\begin{table}[t]
\centering
\resizebox{0.80\textwidth}{!}{%
\begin{tabular}{llccccc}
\toprule
Model & Variant & Params & Extra Params & Train Time (s/epoch) & GPU (GB) & Infer Time (ms) \\
\midrule
\multirow{2}{*}{Matformer}
 & Vanilla        & 3.27M  & ---    & 142.8 & 9.13  & 6.461 \\
 & Prompt          & 3.28M  & 0.01M  & 143.2 & 9.16  & 6.510 \\
\multirow{2}{*}{PotNet}
 & Vanilla        & 0.473M & ---    & 47.0  & 5.82  & 3.060 \\
 & Prompt          & 0.484M & 0.011M & 47.2  & 5.95  & 3.120 \\
\multirow{2}{*}{iComFormer}
 & Vanilla        & 1.38M  & ---    & 72.3  & 9.51  & 4.200 \\
 & Prompt          & 1.39M  & 0.01M  & 72.6  & 9.62  & 4.250 \\
\multirow{2}{*}{eComFormer}
 & Vanilla        & 3.72M  & ---    & 77.8  & 15.90 & 4.580 \\
 & Prompt          & 3.73M  & 0.01M  & 78.0  & 16.10 & 4.600 \\
\bottomrule
\end{tabular}%
}
\caption{Efficiency comparison between vanilla and prompt-infused variants across four backbone GNN models. Prompt learning introduces negligible overhead in parameters, training time, GPU memory, and inference latency.}
\label{tab:efficiency_comparison}
\end{table}
\subsection{Efficiency Study}
\label{time_complex}
We analyze the computational efficiency of different prompt-tuning strategies in terms of the number of trainable parameters, total training FLOPs (measured in GFLOPs), time per epoch, and total training time, as reported in Table~\ref{tab:efficiency}. Compared to SUPT and LEAP, our prompt-based variants of CGCNN and Matformer require substantially fewer training FLOPs while maintaining performance comparable to GPF. In particular, LEAP incurs significantly higher computational cost due to its reinforcement learning–based prompt optimization, resulting in orders-of-magnitude larger training FLOPs and training time. In contrast, our prompt-infused approach introduces only a small number of additional parameters and preserves the lightweight nature of the backbone models, leading to efficient training with minimal computational overhead.\\
To further quantify this overhead relative to the underlying backbones themselves, Table~\ref{tab:efficiency_comparison} reports a direct comparison between the vanilla and prompt-infused variants of four GNN backbones: Matformer, PotNet, iComFormer, and eComFormer -- in terms of parameter count, training time per epoch, peak GPU memory usage, and inference time. The proposed prompts introduce only a negligible number of additional parameters ($\approx$0.01M across all backbones; e.g., Matformer: 3.27M $\rightarrow$ 3.28M), under 0.5\% for larger backbones such as Matformer and eComFormer, and at most 2.3\% for the smallest (PotNet). This overhead is similarly minimal in practice: training time per epoch increases only marginally (typically 0.3--0.7\% across backbones), peak GPU memory usage remains nearly unchanged (0--2\%), and inference time shows negligible overhead (0.4--2\% across backbones). This efficiency stems from the design of the prompt modules, which consist of lightweight linear layers and embedding operations rather than costly additions such as extra message-passing or attention layers, allowing the framework to remain model-agnostic without sacrificing the computational efficiency of the underlying backbone. Overall, the proposed method achieves consistent performance improvements with minimal computational and memory overhead, making it practical for real-world deployment. All experiments were conducted on the JARVIS-DFT dataset using a single NVIDIA A6000 GPU to ensure a fair comparison across methods. This favorable trade-off also holds when viewed analytically rather than just empirically. With $k$ node prompt bases and seven crystal-system graph-level prompts, the total additional parameters amount to $O((k+7)d)$, a modest overhead relative to the multi-layer GNN backbone. Consequently, the proposed method enhances model expressiveness without meaningfully increasing the risk of overfitting.
\begin{table*}[t]
\centering
\begin{minipage}[t]{0.48\textwidth}
\centering
\resizebox{\textwidth}{!}{%
\begin{tabular}{ccccc}
\toprule
Noise & SLME (\%) & Shear Modulus (Gv) & Bulk Modulus (Kv) & Bandgap (MBJ) \\
\midrule
0.00 & 4.764 & 9.581 & 10.050 & 0.285 \\
0.01 & 4.784 & 9.597 & 10.120 & 0.289 \\
0.05 & 4.801 & 9.621 & 10.140 & 0.291 \\
0.10 & 4.812 & 9.654 & 10.190 & 0.295 \\
\bottomrule
\end{tabular}%
}
\caption{Sensitivity of graph-level prompts to noise added to the crystal system label. Noise denotes the label corruption probability.}
\label{tab:noise_sensitivity}
\end{minipage}
\hfill
\begin{minipage}[t]{0.48\textwidth}
\centering
\resizebox{\textwidth}{!}{%
\begin{tabular}{llcccc}
\toprule
Model & Setting & SLME (\%) & Shear Modulus (Gv) & Bulk Modulus (Kv) & Bandgap (MBJ) \\
\midrule
\multirow{2}{*}{Matformer (Only Graph)}
 & OracleSG (Ours) & 4.65 & 9.62 & 10.26 & 0.297 \\
 & UnknownSG        & 4.99 & 9.86 & 10.82 & 0.302 \\
\midrule
\multirow{3}{*}{Matformer (Both)}
 & OracleSG (Ours) & 4.76 & 9.57 & 10.05 & 0.285 \\
 & UnknownSG        & 4.96 & 9.88 & 10.77 & 0.291 \\
& Unsupervised & 4.93 & 9.83 & 10.79 & 0.321 \\
\bottomrule
\end{tabular}%
}
\caption{Effect of space group (SG) availability on graph-level prompting for Matformer.}
\label{tab:oracle_vs_unknown_sg}
\end{minipage}
\end{table*}
\subsection{Robustness to Label Noise and Generalization under Distorted, Label-Free Crystal Inputs}
\label{robust_ablation}
We evaluate robustness under increasing noise levels and in the absence of ground-truth labels, using the Matformer backbone for all experiments in this study. As shown in Table~\ref{tab:noise_sensitivity}, performance degrades only marginally as the noise level increases from 0 to 0.1 (e.g., SLME: 4.764 $\rightarrow$ 4.812), indicating that the model remains stable under input perturbations. To assess label dependence, we further compare against a setting where crystal system labels are unavailable (UnknownSG), which leads to moderate degradation relative to the OracleSG setting (Table~\ref{tab:oracle_vs_unknown_sg}). Introducing a clustering-based unsupervised prompt in place of ground-truth labels recovers most of this gap, achieving performance close to the oracle setting. Together, these results suggest that graph-level prompts are robust to both input distortions and missing supervision, and can be effectively learned in a fully unsupervised manner directly from structural information.
\begin{table}[t]
\centering
\resizebox{0.70\columnwidth}{!}{%
\begin{tabular}{lcccc}
\toprule
Property & GPF & SUPT & LEAP & Prompt (Ours) \\
\midrule
Formation Energy   & 0.032  & 0.031  & 0.039  & \textbf{0.027} \\
Bandgap (OPT)       & 0.135  & 0.139  & 0.134  & \textbf{0.120} \\
Total Energy        & 0.031  & 0.031  & 0.045  & \textbf{0.029} \\
Ehull                & 0.061  & 0.058  & 0.078  & \textbf{0.054} \\
Bandgap (MBJ)       & 0.312  & 0.292  & 0.372  & \textbf{0.287} \\
Bulk Modulus         & 10.69 & 10.66 & 12.13 & \textbf{10.30} \\
Shear Modulus        & 9.659  & 9.558  & 10.35 & \textbf{9.498} \\
SLME                 & 4.822  & 4.909  & 4.789  & \textbf{4.724} \\
Spillage             & 0.362  & 0.359  & 0.383  & \textbf{0.347} \\
\bottomrule
\end{tabular}%
}
\caption{Comparison of graph prompt-tuning baselines (GPF, SUPT, LEAP) against our proposed prompt learning framework on nine properties from the JARVIS-DFT dataset. Best result in each row is highlighted in \textbf{bold}.}
\label{tab:ablation_fairness}
\end{table}
\begin{table}[!htb]
\centering
\setlength{\arrayrulewidth}{0.1mm}
\setlength{\tabcolsep}{7pt}
\renewcommand{\arraystretch}{1.5}
\resizebox{\textwidth}{!}{\begin{tabular}{llcccc}
\toprule
Model & Settings & SLME (\%) & Shear Modulus (Gv) & Bulk Modulus (Kv) & Bandgap (MBJ) \\
\midrule
\multirow{4}{*}{Matformer}
 & node-prompt + Crystal System OneHot   & 4.830 & 9.670 & 10.700 & 0.296 \\
 & node-prompt + Crystal System Embedding & 4.824 & 9.610 & 10.150 & 0.291 \\
 & Conditional Bias                      & 4.851 & 9.961 & 10.845 & 0.301 \\
 & Prompt (Ours)                          & \textbf{4.760} & \textbf{9.581} & \textbf{10.050} & \textbf{0.285} \\
\bottomrule
\end{tabular}}
\caption{Comparison of graph-level conditioning strategies for Matformer on JARVIS-DFT. Best result in each column is highlighted in \textbf{bold}.}
\label{tab:matformer_variants_dis}
\end{table}
 
\subsection{Disentangling the Contribution of Graph-Level Prompts}
\label{graph-prompt}
To further isolate the contribution of graph-level prompts, we compare our method against several baselines that incorporate crystal system information through simpler mechanisms: one-hot concatenation, learned embedding concatenation, and conditional bias (feature-wise affine conditioning via scaling and shifting) ~\cite{dumoulin2018feature-wise}. As shown in Table~\ref{tab:matformer_variants_dis}, access to crystal system information yields only modest gains, and the extent of improvement depends strongly on the integration mechanism. Embedding-based concatenation slightly outperforms one-hot encoding, while the conditional bias baseline fails to improve performance, consistently yielding higher MAE across all evaluated properties. These results suggest that the observed gains are not simply a consequence of exposing the model to crystal system information, but instead depend critically on how that information is integrated, supporting the effectiveness of our proposed graph-level prompting mechanism.
\subsection{Fairness of Baseline Comparison}
\label{fairness_comparison}
In the main paper, baseline results for graph prompt-tuning methods (GPF, SUPT, LEAP) are reported using the settings and hyperparameters published in their original papers. This is standard practice in the graph prompt-tuning literature, since each baseline's hyperparameters, training schedule, and implementation-specific details have already been optimized and validated by its original authors through peer review. Forcing all baselines through a single, uniform pipeline can instead introduce its own sources of unfairness, potentially degrading a baseline's performance relative to what its authors demonstrated and misrepresenting its true capability. To further validate our comparisons, we additionally conduct two controlled experiments that place all baselines under matched training conditions. First, we re-implement GPF, SUPT, and LEAP from scratch on a unified backbone (Matformer), training all methods under identical protocols (optimizer, learning rate schedule, number of epochs, and data splits). Second, to further isolate the effect of pretraining, we fine-tune all methods on top of the same DPF pre-trained Matformer checkpoint, again under identical fine-tuning settings. As shown in Table~\ref{tab:ablation_fairness}, unifying the training protocol modestly improves GPF, SUPT, and LEAP relative to their originally reported numbers, but our method continues to consistently outperform all three across every evaluated property in both the from-scratch and DPF fine-tuning settings. This indicates that our gains stem from the proposed prompt design itself, rather than from differences in backbone, training strategy, or pretraining. We attribute the remaining gap to these baselines reliance on node-centric prompting mechanisms: GPF operates at the node-level, SUPT at the subgraph level, and LEAP performs reinforcement-learning-based prompt editing -- none of which explicitly model crystal-level structural symmetry. In contrast, CrysPrompt incorporates both node-level and crystal-system-aware graph-level prompts, jointly capturing local chemical semantics and global structural symmetry. This multi-level design yields consistent improvements even under matched training conditions, with average gains of 7.12\% over GPF, 5.85\% over SUPT, and 18.27\% over LEAP on JARVIS-DFT, demonstrating the effectiveness of our proposed multi-level prompt design.

\begin{figure}[!htb]%
    \centering
    {\includegraphics[width=\textwidth]{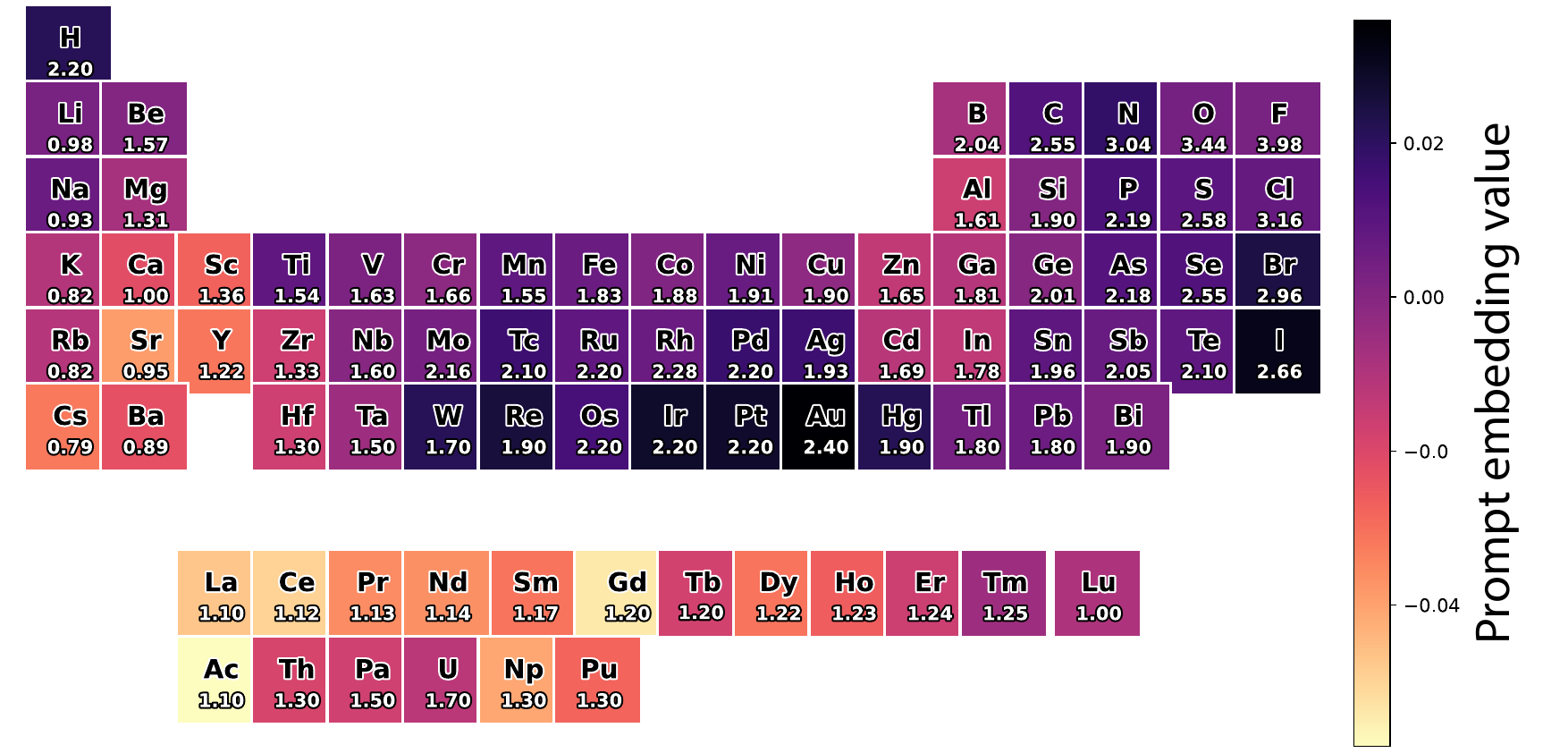}}
    {\caption{Periodic table visualization of the mean node-level prompt embedding values for dimension 84 across elements. Tile colors denote the prompt embedding values, while the numbers inside each tile represent the electronegativity of the corresponding element (Pauling scale).}
    \label{fig:periodic_crystals_en}}
\end{figure}

\begin{figure}[!htb]%
    \centering
    {\includegraphics[width=\columnwidth]{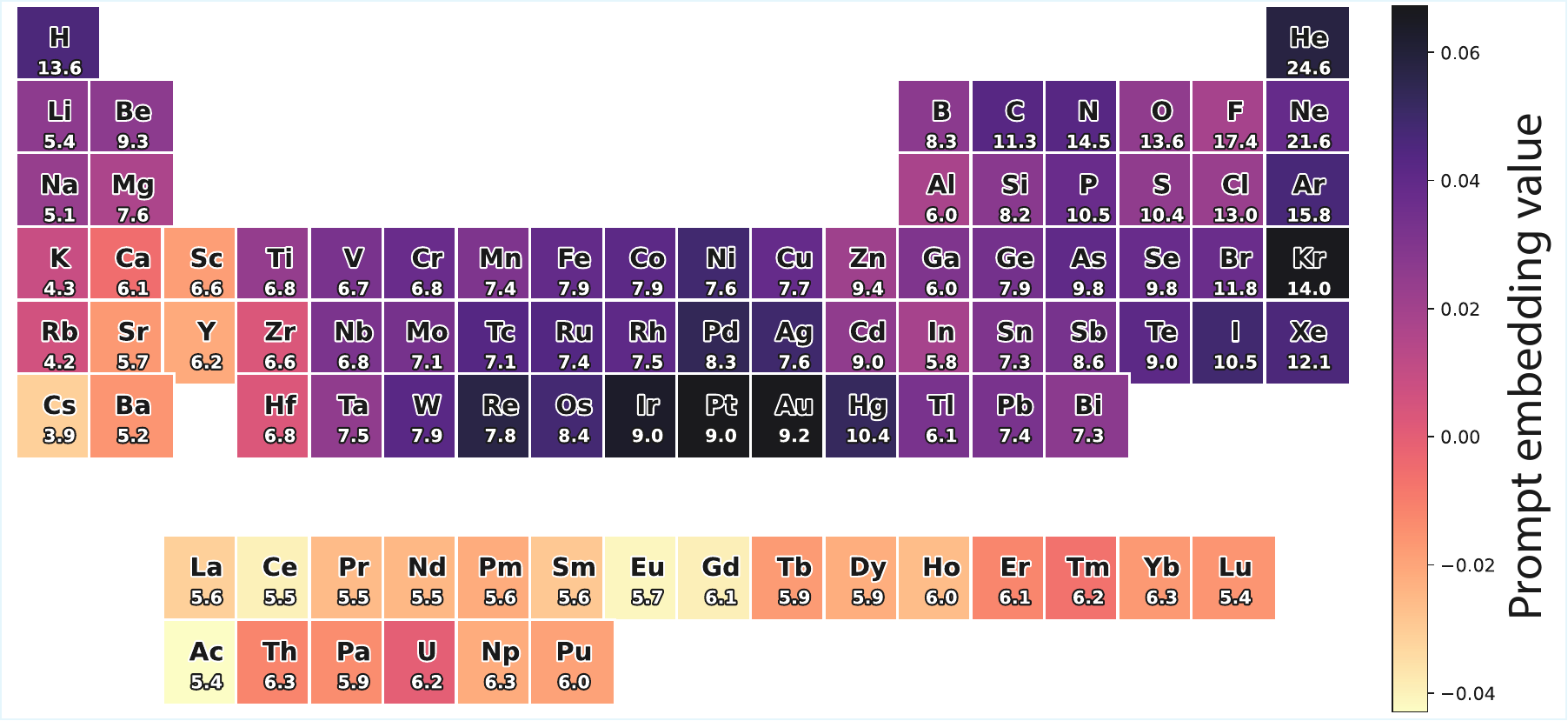}}
    {\caption{Periodic table visualization of the mean node-level prompt embedding values for dimension 73 across elements. Tile colors represent prompt values, while the numbers denote the first ionization energy (eV) of each element. The learned prompt dimension exhibits periodic trends consistent with the variation of ionization energy across the periodic table.}
    \label{fig:periodic_crystals_ion}}
\end{figure}

\subsection{Expressiveness of Node-Level Prompt}
\label{expressiveness_node}
To further analyze the expressiveness of the learned node-level prompts, we investigate whether individual prompt dimensions capture meaningful chemical trends across elements. For each element, we compute the mean value of the learned prompt embeddings across all occurrences of that element in the dataset. These aggregated values are then visualized over the periodic table to examine whether the learned prompt representations exhibit patterns consistent with known chemical properties.
\\
Figs.~\ref{fig:periodic_crystals_en} and~\ref{fig:periodic_crystals_ion} illustrate representative prompt dimensions that correlate with fundamental elemental descriptors. In Figure 5, the tile colors represent the mean prompt embedding values for dimension 84, while the numbers inside each tile denote the Pauling electronegativity of the corresponding element. The learned prompt values follow periodic patterns consistent with electronegativity variations across the periodic table. Elements with higher electronegativity—such as O, F, Cl, and Br—tend to exhibit larger prompt values, while elements with lower electronegativity, particularly heavier electropositive metals (e.g., Cs and Ba), generally correspond to comparatively smaller prompt values~\cite{pauling1932nature}. Quantitatively, we observe a strong Pearson correlation of 0.784 between the prompt values and electronegativity.
\\
As illustrated in Fig.~\ref{fig:periodic_crystals_ion}, prompt dimension 73 exhibits periodic patterns consistent with the variation of first ionization energy across the periodic table. Electropositive metals with low ionization energies, such as Cs, Rb, and K, tend to correspond to smaller prompt values (lighter colors). In contrast, transition metals with comparatively higher ionization energies, such as Pt, Au, and Hg, exhibit larger prompt values (darker colors)~\cite{mulliken1934new}. Quantitatively, we observe a Pearson correlation coefficient of 
$r=$0.742 between the prompt values and the ionization energy, indicating that the learned prompt dimension captures chemically significant trends related to the atomic electron binding strength.
\begin{figure}[!htb]%
    \centering
{\includegraphics[width=0.80\columnwidth]{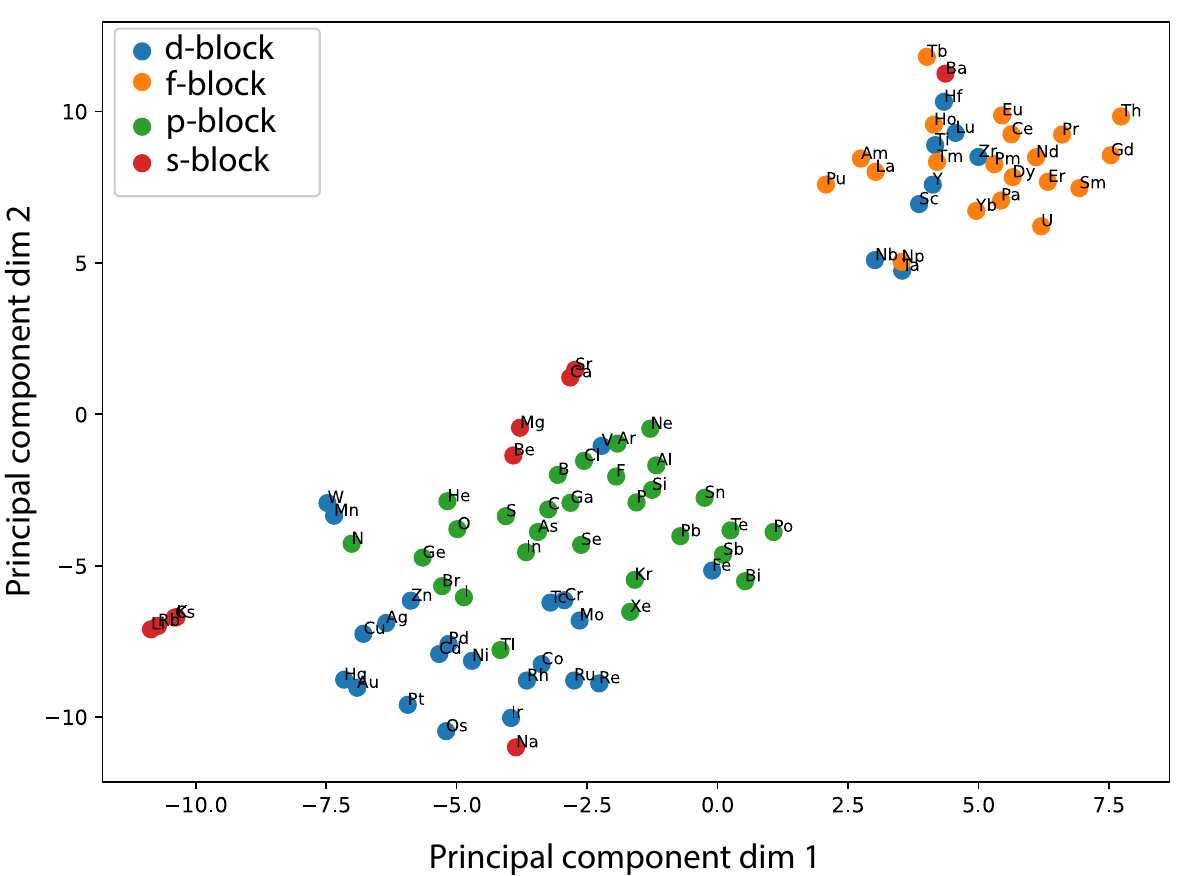}}
    {\caption{Visualization of node-level prompt embeddings across periodic blocks. Two-dimensional t-SNE projection of node-level prompt embeddings. Each point represents a chemical element and is colored by its periodic block (s, p, d, or f).}
    \label{fig:node_prompt_interpret}}
\end{figure}

\begin{figure}[!htb]%
    \centering
    {\includegraphics[width=0.6\columnwidth]{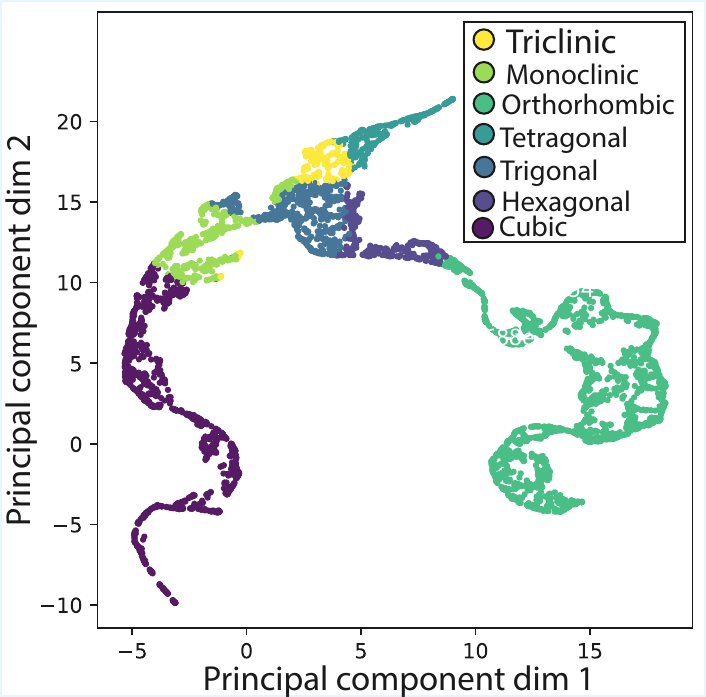}}
    {\caption{Visualization of graph prompt embeddings across crystal systems. Two-dimensional projection of graph-level prompt embeddings using principal component analysis (PCA). Each point represents a crystal structure and is colored according to its crystal systems.}
    \label{fig:graph_prompt}}
\end{figure}

\subsection{Expressiveness of Graph-Level Prompt}
\label{expressiveness_graph}
To evaluate whether the learned graph-level prompts capture meaningful structural information, we analyze the prompt embeddings using unsupervised clustering. Specifically, we apply K-means clustering with $k=$7, corresponding to the seven crystal systems, on the learned graph prompt embeddings and visualize them using Principal Component Analysis~\cite{abdi2010principal}. As shown in Fig.~\ref{fig:graph_prompt}, the embeddings form well-separated clusters that align with crystallographic symmetry groups. Quantitatively, the clustering achieves Normalized Mutual Information (NMI)=0.70, Adjusted Rand Index (ARI)=0.75, and a Silhouette score of 0.642, indicating that the learned prompt embeddings capture significant structural information about crystal symmetry. The relatively high agreement between unsupervised clusters and ground-truth crystal system labels suggests that the graph-level prompts encode meaningful global geometric characteristics of crystalline materials. Overall, this analysis demonstrates that the learned prompts are expressive and capture structural patterns consistent with crystallographic symmetry. 

\end{document}